\documentclass[pubjcg,nonpub]{preprint}
\setcopyright{none}

\usepackage{booktabs}
\usepackage{multirow}
\usepackage{subcaption}
\usepackage{pgfplots}
\usepackage{siunitx}
\pgfplotsset{compat=1.18}
\usepackage{makecell}

\usepackage[ruled]{algorithm2e}

\SetAlFnt{\small}
\SetAlCapFnt{\small}
\SetAlCapNameFnt{\small}
\SetAlCapHSkip{0pt}

\newcommand{\figref}[1]{Fig.~\ref{#1}}
\newcommand{\secref}[1]{Sec.~\ref{#1}}

\newcommand{\eqnref}[1]{Eq.~\eqref{#1}}
\newcommand{\tabref}[1]{Tab.~\ref{#1}}

\newcommand{\method}{LSO\xspace}

\begin{document}

\title{BFMTrack: Latent Sequence Optimization for Physics-Based Motion Tracking with Behavioral Foundation Models}

\author{Thomas Rupf}
\email{thomas.rupf@disneyresearch.com}
\orcid{0009-0001-5971-3115}
\affiliation{%
  \institution{Disney Research}
  \city{Zurich}
  \country{Switzerland}
}

\author{Agon Serifi}
\email{agon.serifi@disneyresearch.com}
\orcid{0000-0003-4439-0023}
\affiliation{%
  \institution{Disney Research}
  \city{Zurich}
  \country{Switzerland}
}

\author{David M{\"u}ller}
\email{david.mueller@disneyresearch.com}
\orcid{0009-0001-6591-8803}
\affiliation{%
  \institution{Disney Research}
  \city{Zurich}
  \country{Switzerland}
}

\author{Sammy Christen}
\email{sammy.christen@disneyresearch.com}
\orcid{0000-0002-3511-8565}
\affiliation{%
  \institution{Disney Research}
  \city{Zurich}
  \country{Switzerland}
}

\author{Ruben Grandia}
\email{ruben.grandia@disneyresearch.com}
\orcid{0000-0002-8971-6843}
\affiliation{%
  \institution{Disney Research}
  \city{Zurich}
  \country{Switzerland}
}

\author{Espen Knoop}
\email{espen.knoop@disneyresearch.com}
\orcid{0000-0002-7440-5655}
\affiliation{%
  \institution{Disney Research}
  \city{Zurich}
  \country{Switzerland}
}

\author{Moritz B{\"a}cher}
\email{moritz.baecher@disneyresearch.com}
\orcid{0000-0002-1952-1266}
\affiliation{%
  \institution{Disney Research}
  \city{Zurich}
  \country{Switzerland}
}

\begin{abstract}
 Behavioral Foundation Models (BFMs) offer a promising path toward universal physics-based character control by organizing a rich repertoire of physically plausible behaviors into a latent space, guided by a large-scale motion dataset. While these models excel at time-invariant tasks, such as goal-reaching and state-based reward optimization, their latent space does not directly support time-varying objectives, such as tracking a motion sequence. For tracking, existing heuristics rely on moving-window-averaging that fails to capture the nuances of highly dynamic motions. In this work, we propose a novel Latent Sequence Optimization (\method) to address these shortcomings.
 Our approach combines simulation rollouts with a policy gradient update to optimize over a sequence of latents, extending the capabilities of BFMs toward precise motion tracking without requiring reward engineering and tuning. To guide the optimization toward smooth, coherent latent trajectories, we model the latent sequence using temporally correlated noise. We validate our approach across dense tracking, sparse keyframing, and direct deployment onto a real humanoid robot.
\end{abstract}

\keywords{robotic character control, motion tracking, physics-based characters, robotics}

\begin{teaserfigure}
    \centering
    \includegraphics[width=\linewidth,trim={24cm 27cm 19cm 30cm},clip]{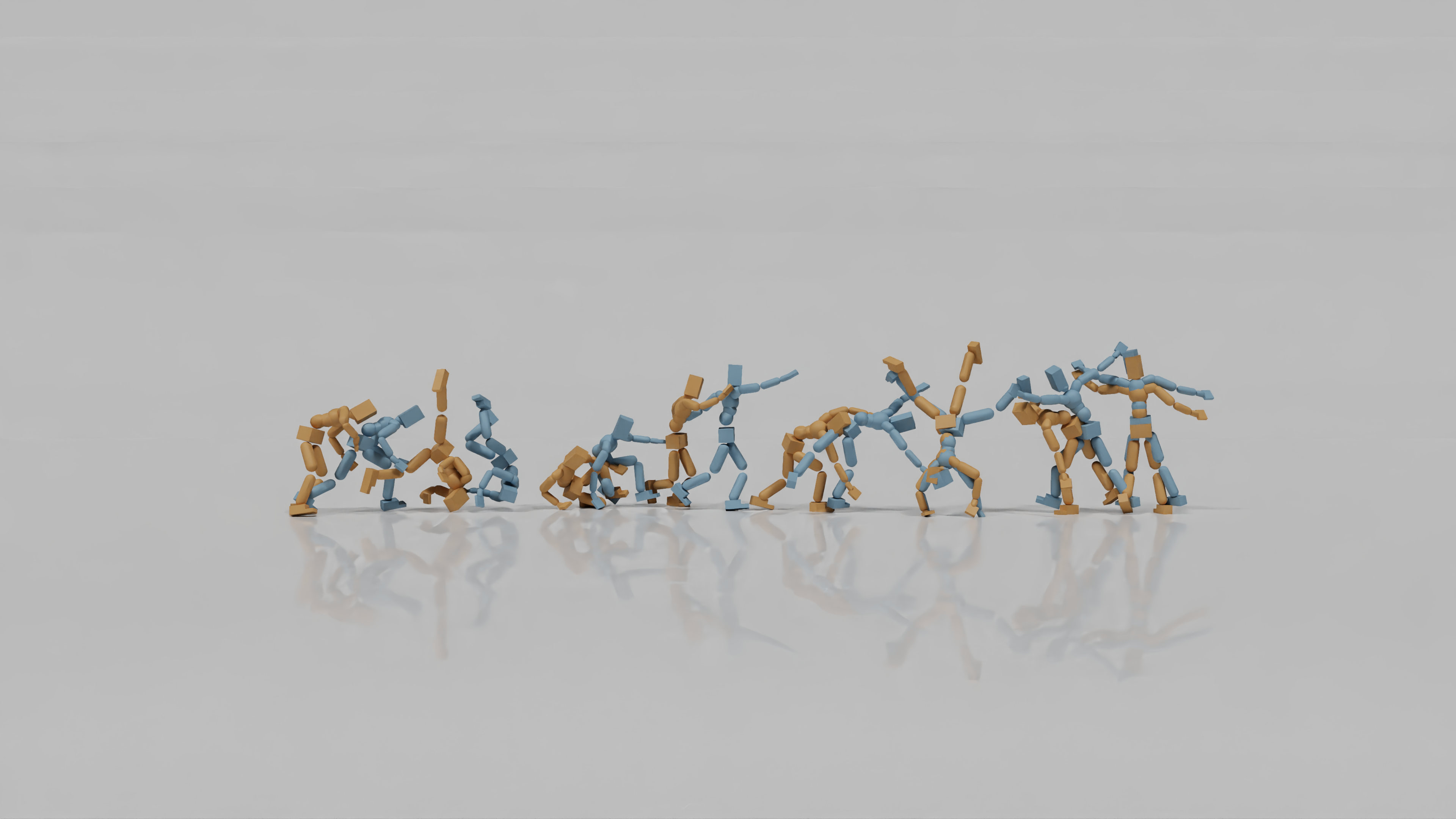}
    \caption{Our method repurposes a trained Behavioral Foundation Model (BFM) for motion tracking by optimizing a temporal sequence of latents to approximate a dense kinematic reference or a sparse set of  keyframes (in blue) with physics-aware motion (in yellow).}
    \Description{A yellow humanoid shown at multiple instances in time hitting different keyframes shown as a blue humanoid.}
  \label{fig:teaser}
\end{teaserfigure}

\maketitle

\section{Introduction}
\label{sec:intro}

Behavioral Foundation Models (BFMs) have recently been proposed for learning universal control policies for physics-based characters using reward-free interactions with the environment~\cite{touatiLearningOneRepresentation2021,touatiDoesZeroShotReinforcement2023}. Guided by large-scale datasets of reference motions~\cite{tirinzoni2025zero, li2025bfm}, they provide a leap forward towards the goal of training controllable characters from \emph{unsupervised} interactions: acting with a policy that is optimal for a reward that is specified after training, without requiring retraining. Because BFMs encode the solutions of \emph{all} possible interactions with the environment, there is practically no undefined behavior, resulting in remarkably stable and versatile physics-based character performance. However, so far, the demonstrated capabilities have mostly been limited to stationary tasks: optimizing reward functions, such as a forward velocity to make a character walk forward or reaching single-frame target poses.  

A natural question is how far these capabilities extend to \emph{motion tracking}, i.e., reproducing a time-varying reference trajectory with a physics-based character. Motion tracking is inherently non-stationary: the desired behavior changes at every timestep, making it challenging to design an accurate reward function that tracks a desired reference motion and maps to the correct latent. Current BFMs address the non-stationarity of tracking by compressing reference segments into single latents through a moving-window heuristic \cite{tirinzoni2025zero, li2025bfm}: a window of several goal states is summarized into one latent vector via \emph{permutation-invariant} aggregation, discarding temporal order within the look-ahead window. 

While surprisingly effective for many motions, this heuristic imposes a fundamental coherence-precision trade-off: large windows provide motion coherence when the next segment is consistent, but can wash out fast transitions; small windows preserve local precision but lack the context needed for anticipatory movements such as a preparatory step before a kick. Moreover, BFMs are trained without latent switches, leading to suboptimal behavior when concatenating latents naively during inference.

In this work, we propose Latent Sequence Optimization (\method) to \emph{optimize a temporal sequence of latents}, moving away from instantaneous latent selection. Given a reference motion, our approach directly optimizes a latent trajectory to reproduce the reference by combining simulator rollouts with a policy gradient update. Notably, the optimization loop is self-contained, and no additional reward tuning is required. Our optimization accounts for global coherence of motions while maintaining local precision, resolving the previously mentioned trade-off. 

Beyond tracking of dense motion sequences, our method naturally extends to ``inbetweening'' for sparse keyframes, producing expressive, full-body motions grounded in physics and in the motion space of the training data. This capability distinguishes our approach from purely kinematic methods, which lack physical guarantees, and from dedicated tracking policies, which require dense reference trajectories.

We validate our framework on a simulated SMPL humanoid and demonstrate its applicability on a humanoid robot. Our latent sequence optimization consistently outperforms zero-shot BFM inference across all evaluated metrics. Beyond dense tracking, we show that our framework extends to sparse keyframe inputs, enabling tracking of under-specified, highly dynamic, contact-rich motions.

Succinctly, our main contributions are:
\begin{itemize}
    \item A latent optimization method for BFM motion tracking, outperforming baseline methods while requiring no manual reward engineering. %
    \item Generalization of the latent optimization to sparse keyframe tracking via a temporal regularization on the latent sequence.
    \item Demonstrations of the framework's versatility across dense tracking, sparse keyframes, and motion stitching, validated in simulation on a SMPL humanoid and with a sim-to-real deployment on a humanoid robot.
\end{itemize}

\section{Related Work}

\paragraph{Physics-Based Motion Tracking}

Tracking reference motions on physics-based characters is a long-standing topic in animation and robotics. The canonical per-clip approach, DeepMimic~\cite{pengDeepMimicExampleGuidedDeep2018}, trains a phase-synchronized RL policy with hand-crafted tracking rewards. This tight clip-policy coupling implicitly captures reference dynamics for anticipatory control, but requires retraining per clip. Subsequent works scale the number of motions a tracker can imitate via multiple experts~\cite{won2020scalable, huang2025modskill}, model-based components~\cite{fussell2021supertrack, yao2024moconvq, li2025robotic}, learned motion priors~\cite{luo_perpetual_2023, won2022physics, serifi2024vmp}, diffusion-based policies~\cite{liao2025beyondmimic, huang2025diffusecloc, truong2024pdp, wu2025uniphys}, or large-scale training~\cite{luo2025sonic,tessler2024maskedmimic}. However, tracking remains the central supervision signal: when the reference is infeasible or out of sync with the current state (e.g., after a disturbance), policies tend to overcorrect, leading to jitter and unnatural behavior.

Adversarial methods replace explicit motion-tracking rewards with a discriminator that implicitly learns a natural motion distribution~\cite{pengAMPAdversarialMotion2021,xu2021gan}, loosening the strict frame-by-frame supervision. However, adversarial training can be unstable and prone to mode collapse, resulting in a reduced skill set. 

An orthogonal challenge is motion tracking of sparse keyframes \cite{agrawal2024skel, harvey2020robust}. This is difficult because supervision signals are sparse and the inter-keyframe motion is under-determined. Existing approaches train policies that bake the sparse-keyframe interface into the controller via specialized reward decompositions~\cite{zargarbashi2025robotkeyframing} or staged adaptation modules~\cite{huang2025adaptive}. 

We take another route for motion tracking by repurposing BFMs, which approximately encode the solutions of \emph{all} interactions within an environment, avoiding common jittering artifacts and resulting in remarkably stable motions on hardware. By optimizing latent sequences, we not only show that the space can serve dense motion tracking tasks, but also provides a good motion prior for sparse keyframe regimes, without requiring reward tuning or specialized training methods.

\paragraph{Unsupervised Reinforcement Learning} Un- and self-supervised RL aim to learn foundational policy families from reward-free interactions that can be prompted with new rewards at inference time~\cite{touatiLearningOneRepresentation2021,parkFoundationPoliciesHilbert2024,bagatellaTDJEPALatentpredictiveRepresentations2025,jajooRegularizedLatentDynamics2025,zhengCanMISLFly2025}. The Forward-Backward (FB) framework~\cite{touatiLearningOneRepresentation2021} results in a reward-centric latent space that serves as a principled prompt space for latent optimization~\cite{touatiDoesZeroShotReinforcement2023}, unlike constrained RL~\cite{hugessenZeroShotConstraintSatisfaction2025} and imitation learning~\cite{pirottaFastImitationBehavior2023} formulations.  
FB-CPR~\cite{tirinzoni2025zero} extends the FB framework with a motion-capture-conditioned discriminator that regularizes the learned policy family towards human-like behaviors. By using auxiliary rewards, this formulation has been successfully applied to a humanoid robot~\cite{li2025bfm}.

FB-based policies are smooth w.r.t.~latents \cite{bagatella2026softforwardbackwardrepresentationszeroshot}, and recent latent-space adaption methods refine them at test time, but remain limited in scope: LoLA~\cite{sikchiFastAdaptationBehavioral2025} runs policy gradient on a single latent for stationary reward optimization, OpTI-BFM~\cite{rupfOptimisticTaskInference2025} actively infers a single reward latent through online interaction, and BFM-Zero~\cite{li2025bfm} performs zero-order optimization over a single latent (or a short window) for few-shot adaptation under altered model dynamics such as increased mass or reduced friction. 

In contrast, we propose latent-sequence optimization on the full latent trajectory using policy gradients, outperforming related approaches for tracking, while also enabling sparse-keyframe-based control, which is not possible with other techniques. 

Conceptually adjacent, mutual information skill learning methods \cite{pengASELargeScaleReusable2022,cathomenDivideDiscoverDeploy2025} or skill-embedding methods, like CALM \cite{tesslerCALMConditionalAdversarial2023}, also extract a continuous latent skill space from motion data. However, they often fail to recover the full range of optimal policies \cite{eysenbachInformationGeometryUnsupervised2021}. The reward-centric approach in BFMs, on the other hand, yields superior task performance and motion diversity \cite{tirinzoni2025zero}, making BFMs a compelling foundation for motion tracking.

\section{Overview}

\begin{figure}
    \centering
    \includegraphics{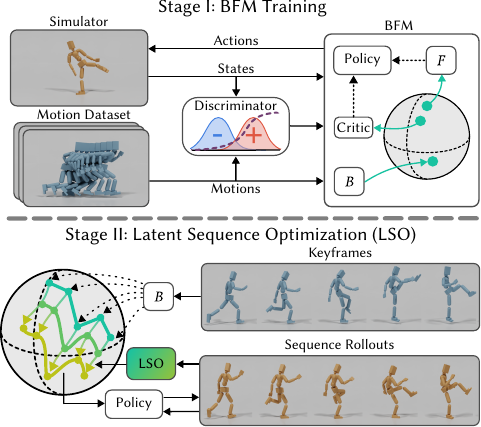}
    \caption{\textbf{Overview.} In Stage I, we train a latent-conditioned policy via the Forward-Backward framework. In Stage II, we map the goal keyframes into the latent space via the backward map $B$, further optimizing the latent sequence to improve the temporal coherence and tracking accuracy.}
    \label{fig:overview}
\end{figure}

\paragraph{Problem Statement}
We focus on tracking reference motion on physics-based characters using a control policy. More specifically, we consider a sequence of character states $\{\mathbf{s}_t\}_{t=1}^{T}$ and a sequence of goal keyframes $\{ (\mathbf{g}_{t_k}, t_k) \}_{k=1}^{K}$, where $t_k$ represents the timestep of the $k$-th keyframe. 
The density of the keyframes is flexible: $K=1$ indicates a single goal keyframe and $K=T$ represents a dense set of keyframes. 
Our goal is to find a policy $\pi$, conditioned on the current state $\mathbf{s}_t$, that produces actions $\mathbf{a}_t$ that enable accurate tracking of the keyframes, and, in a sparse regime, infills plausible physics-informed motions between consecutive keyframes. 

\paragraph{Solution} Our approach is illustrated in \figref{fig:overview} and proceeds in two stages. Stage I trains a latent-conditioned policy $\pi(\mathbf{s},\mathbf{z})$ via the Forward-Backward (FB) framework (\secref{sec:fb}). Stage II uses the framework to embed the time-varying goals as a sequence of latents, yielding an initial but non-optimal motion. We then optimize this latent sequence $\{\mathbf{z}_{t}\}_{t=1}^{T}$ (\secref{sec:latent_optimize}), asking it to optimally track the reference motion. For sparse keyframes, we exploit the smoothness of the latent space to fill in the gaps with plausible motion (\secref{sec:inductive_biases}). 

\section{FB Framework: Background and Intuition}
\label{sec:fb}

To learn a diverse set of behaviors for a physics-based character, we build on the FB framework~\cite{touatiLearningOneRepresentation2021} to train the latent-conditioned policy $\pi(\mathbf{s},\mathbf{z})$, where the latent code $\mathbf{z} \in \mathcal{Z}$ lies on the unit hypersphere in $\mathbb{R}^{d}$. The FB framework assumes that the policy operates on a reward-free Markov decision process (MDP) and is trained using unsupervised interactions with the environment. Two auxiliary networks are introduced to train the policy: the \emph{backward map} $B: \mathbf{s} \mapsto B(\mathbf{s}) \in \mathcal{Z}$, which embeds each state into the latent space, and the \emph{forward map} $F: (\mathbf{s}, \mathbf{a}, \mathbf{z}) \mapsto F(\mathbf{s}, \mathbf{a}, \mathbf{z}) \in \mathbb{R}^{d}$, which predicts the ``value'' of an action $\mathbf{a}$ to achieve the encoded goal $\mathbf{z}$ when starting from $\mathbf{s}$ and following the policy. 
Once trained, the BFM framework enables zero-shot mapping from any reward function $r(\mathbf{s})$ to an optimal policy. This is achieved by computing a reward-weighted sum of backward embeddings $\mathbf{z} = \sum_i B(\mathbf{s}_i) \cdot r(\mathbf{s}_i)$ over the states visited during training. The policy $\pi(\mathbf{s}, \mathbf{z})$ is then conditioned on this latent vector. Notably, when the goal is to reach a specific target state, defined by an indicator reward, this expression simplifies to $\mathbf{z} = B(\mathbf{s})$.

\paragraph{Training} Training is performed off-policy using collected transitions $(\mathbf{s}_t, \mathbf{a}_t, \mathbf{s}_{t+1})$, jointly optimizing the forward and backward maps, such that $\pi(\mathbf{s}, \mathbf{z}) := \arg \max_{\mathbf{a}} F(\mathbf{s}, \mathbf{a}, \mathbf{z})^T \mathbf{z}$ and $F(\mathbf{s}, \mathbf{a}, \mathbf{z})^T B(\mathbf{s}')$ approximates the probability to reach any $\mathbf{s}'$ from $\mathbf{s}$ if following $\pi$. Each transition is replayed with many different latent codes, allowing $\pi$ to learn how to realize any latent from any state~\cite{touatiLearningOneRepresentation2021}. 

To make FB tractable for humanoid control, we rely on two extensions for training: first, we implement conditional policy regularization (FB-CPR) \cite{tirinzoni2025zero}, which introduces a latent-conditioned discriminator that incentivizes FB to discover behaviors that stay close to the behaviors found in a large-scale human motion dataset. Second, following \cite{li2025bfm}, we use reward terms, such as action rate and torque regularization, which facilitate the sim-to-real transfer. The additional reward feedback from the discriminator and the simulator is integrated with FB by training a \emph{critic network} $Q: (\mathbf{s}, \mathbf{a}, \mathbf{z}) \mapsto Q(\mathbf{s}, \mathbf{a}, \mathbf{z}) \in \mathbb{R}$. To enable inference, we also explicitly train the actor network $\pi(\mathbf{s}, \mathbf{z})$.

\paragraph{Inference}
In the dense tracking problem ($K = T$), where target states are prescribed for every timestep, existing BFM-based methods~\cite{tirinzoni2025zero,li2025bfm} assume that the motion over a horizon $L$ can be approximated by a stationary behavior found by inferring the empirical reward (ER) imitation latent~\cite{pirottaFastImitationBehavior2023}
\begin{align}
    \mathbf{z}_t = \frac{1}{L} \sum_{\tau=t+1}^{t+L} B(\mathbf{g}_\tau).
    \label{eq:zt-lookahead}
\end{align}

This aggregation is \emph{permutation-invariant}, because it discards the temporal order of states, resulting in the aforementioned coherence-precision trade-off. In addition, the per-timestep latent switching falls outside the policy's training distribution. Unlike dedicated tracking methods~\cite{pengDeepMimicExampleGuidedDeep2018,luo_perpetual_2023,serifi2024vmp}, BFMs do not explicitly model the passive dynamics of a non-stationary reference.

\section{BFMs for Motion Tracking}

While zero-shot tracking with BFMs inherently faces a coherence-precision trade-off, decoupling the policy from temporal dynamics provides a crucial advantage: it enables broader applications, such as sparse keyframe tracking, without requiring the retraining of a specialized tracking policy. To retain this flexibility while overcoming the drawbacks of ER-based rewards, we propose Latent Sequence Optimization (\method) to directly optimize the entire latent sequence $\{\mathbf{z}_t\}_{t=1}^T$ using simulation rollouts.

\subsection{Dense Motion Tracking via Latent Optimization}
\label{sec:latent_optimize}

Gradient-based trajectory optimization is not an option in our context, because we rely on simulation rollouts for future states, making state-dependent objectives non-differentiable. Single-latent adaptation has been demonstrated for BFMs~\cite{sikchiFastAdaptationBehavioral2025} and we extend this to non-stationary motion tracking by treating the latent sequence as a continuous stochastic optimization problem. 

For a target motion of length $T$, we define Gaussian distributions $\mathbf{z}_t \sim \mathcal{N}\!(\boldsymbol{\mu}_t, \sigma^2 \mathbf{I})$ with fixed standard deviation $\sigma$, where the per-timestep means $\{\boldsymbol{\mu}_t\}_{t=1}^{T}$ serve as optimization variables. Leveraging the backward map, we initialize the sequence of means with latents that encode the task of reaching the target pose at each timestep, i.e. $\boldsymbol{\mu}_t = B(\mathbf{g}_t)$.
We then refine the latent sequence by using a REINFORCE-style policy gradient update~\cite{williams1992reinforce}. To this end, for each optimization step, we sample $N$ latent sequences of length $T$
\begin{align}
    \mathbf{z}_t^{(n)} \sim \mathcal{N}\!(\boldsymbol{\mu}_t, \sigma^2 \mathbf{I}) \quad \text{for } t=1,\dots,T \text{ and } n=1,\dots,N,
\end{align}
and roll them out in $N$ parallel simulation environments, observing the resulting states $\mathbf{s}_t^{(n)}$. The tracking reward $r_t^{(n)}$ is computed directly in the latent space using the cosine similarity between the backward embeddings of the simulated state and the target goal
\begin{align}
    r_t^{(n)} = \frac{B(\mathbf{s}_t^{(n)})^T B(\mathbf{g}_t)} {\|B(\mathbf{s}_t^{(n)})\|_2 \, \|B(\mathbf{g}_t)\|_2}.
\end{align}
This makes the optimization loop self-contained: the backward map provides both the initialization of the latent sequence and the reward signal, eliminating the need for additional reward design. Note that this framework remains compatible with arbitrary reward functions defined over the simulated state $\mathbf{s}_t$ or latent vectors $\mathbf{z}_t$, enabling extensions that include additional objectives such as velocity targets or desired contact patterns.

From these timestep rewards, we compute the discounted returns $R_t^{(n)} = \sum_{j=t}^{T} \gamma^{j-t} r_j^{(n)}$ with discount factor $\gamma$. 
Using the leave-one-out baseline \cite{kool2019buy} to compute an advantage estimate with reduced variance
\begin{align}
    \overline{R}_t^{(n)} = \left(R_t^{(n)} - \frac{1}{N-1}\sum_{i=1, i\ne n}^N R_t^{(i)}\right),
\end{align}
we update our mean latent trajectory at each timestep $t$ along the direction
\begin{align}
    \Delta \boldsymbol{\mu}_t = \frac{1}{N}\sum_{n=1}^N \overline{R}_t^{(n)} \nabla_{\boldsymbol{\mu}_t} \log \mathcal{N}\!(\mathbf{z}_t^{(n)} | \boldsymbol{\mu}_t, \sigma^2 \mathbf{I}).
\end{align}
Because the latent space is restricted to the unit hypersphere, we maintain geometric feasibility by projecting the updated mean parameters back onto the sphere via $L_2$ normalization after each optimization step
\begin{align}
    \boldsymbol{\mu}_t \leftarrow \frac{\boldsymbol{\mu}_t}{\|\boldsymbol{\mu}_t\|_2}.
\end{align}

Sampling latent trajectories with independent per-step \emph{white} noise is suboptimal: because each latent vector encodes a distinct behavior, uncorrelated perturbations at every timestep cause frequent transitions between unrelated motions, reducing the overall coherence of the rollouts. Furthermore, these high-frequency variations tend to average out over the trajectory. Consequently, the sampled sequences remain close to their mean, yielding similar returns across different rollouts and attenuating the policy gradient signal. We instead bias the exploration toward smooth, \emph{coherent} latent trajectories by sampling the full sequence as temporally correlated \emph{colored noise}~\cite{pinneriSampleefficientCrossEntropyMethod2020}, with power spectral density $S(f) \propto 1/f^\beta$~\cite{timmer1995generating}. We nominally set $\beta=1$ (\emph{pink noise}) as this has been proven effective for policy gradient-based methods~\cite{eberhard2023pink}. After optimization, the final latent sequence is set to the optimized means: $\{\boldsymbol{z}_t\}_{t=1}^{T} = \{\boldsymbol{\mu}_t\}_{t=1}^{T}$.

\subsection{Sparse Keyframe Tracking}
\label{sec:inductive_biases}

The dense tracking formulation extends naturally to the sparse setting, where goals are specified at a sparse set of keyframes. Because no reference is available between these points, we set the tracking reward $r_t = 0$ at all intermediate timesteps. We begin the optimization process by constructing a dense sequence of latent means $\{\boldsymbol{\mu}_t\}_{t=1}^T$, initializing the latents with zero-shot encodings $B(\mathbf{g}_{t_k})$ at keyframes and interpolating the remaining timesteps via Spherical Linear Interpolation (SLERP)~\cite{shoemake1985animating}. Finally, we optimize the entire latent sequence using the policy gradient method introduced in~\secref{sec:latent_optimize}.

Correlating the Gaussian sampling noise naturally imposes a crucial structural prior on the optimization of the latent sequence. As keyframe sparsity increases, the tracking objective becomes highly under-determined, as any trajectory passing through the keyframes is a valid solution. The resulting latent sequence must not only hit the keyframes but also synthesize plausible motion between them. Therefore, imposing structural priors on the latent sequence is essential to effectively regularize the optimization space of the policy gradient formulation. In our experiments (see \secref{sec:eval_sparse} and supplemental video), we ablate this prior against other valid methods for imposing structure, such as holding the latent constant between keyframes or parameterizing the trajectory with a continuous spherical spline.

\section{Implementation Details}
\label{sec:implementation_details}

We evaluate our framework on two embodiments using Isaac Sim \cite{makoviychuk2021isaac} as the simulator: a 69-DoF SMPL humanoid \cite{yuan2020residual, loper2015smpl} and Lima, a custom 20-DoF bipedal robot that we additionally use for real-world deployment. For both platforms, the actions of the control policy are position targets for proportional-derivative (PD) controllers. Because off-policy training methods require bounded action spaces for improved training convergence, we restrict the policy outputs to $[{-1}, 1]$ and rescale them to the full joint ranges with a margin, enabling the PD controller to deliberately overshoot to exert torque near the joint limits. 

While this general control scheme applies to both embodiments, specific simulation and network parameters differ. The humanoid utilizes an idealized actuator model \cite{serifi2024vmp} running at \SI{30}{\hertz} with a 6-layer residual network. Lima uses more realistic actuator models~\cite{grandia2024}, runs at \SI{50}{\hertz} and uses a smaller 4-layer MLP to meet the real-time constraints of its onboard Intel Core i7 CPU. We summarize key parameters in~\tabref{tab:embodiments}.

\begin{table}[t]
  \centering
  \caption{\textbf{Implementation Overview.} We evaluate our framework on two embodiments: an SMPL humanoid and a custom biped (Lima). While both share the same FB-CPR pre-training recipe, their architectures and simulation parameters differ to accommodate Lima's real-world compute and hardware constraints.}
  \label{tab:embodiments}
  \resizebox{\columnwidth}{!}{%
  \begin{tabular}{@{}lll@{}}
    \toprule
    \textbf{Parameter}            & \textbf{SMPL Humanoid}        & \textbf{Lima Biped} \\
    \midrule
    \multicolumn{3}{@{}l}{\textit{Hardware \& Simulation}} \\
    Degrees of Freedom (DoF)      & 69                            & 20 \\
    Control rate                  & \SI{30}{\hertz}               & \SI{50}{\hertz} \\
    Actuation                     & PD (idealized)                & PD (w/ actuator model) \\
    Domain randomization          & ---                           & Mass, friction, forces, obs.\ noise \\
    \addlinespace
    \multicolumn{3}{@{}l}{\textit{Network Architecture}} \\
    Latent dimension $d$          & 256                           & 256 \\
    Actor-Critic setup            & Symmetric                     & Asymmetric (privileged critic) \\
    Actor $\pi$                   & Residual, $6\!\times\!2048$   & MLP,  $[1024, 1024, 512, 512]$ \\
    Forward map $F$ / critic $Q$  & Residual, $6\!\times\!2048$   & Residual, $6\!\times\!1024$ \\
    Backward map $B$ / disc.     & MLP, $3\!\times\!1024$        & MLP, $3\!\times\!1024$ \\
    \bottomrule
  \end{tabular}%
  }
\end{table}

We define two state representations as input to our networks. The control policy relies on the actor state $\mathbf{s}^\text{act}_t$, while the backward map and the discriminator use the motion state $\mathbf{s}^\text{mot}_t$:
\begin{equation}
\begin{aligned}
    \mathbf{s}^\text{act}_t  &:= (\hat{\mathbf{g}}_t,\, \mathbf{v}_t,\, \boldsymbol{\omega}_t,\, \mathbf{q}_t,\, \dot{\mathbf{q}}_t,\, \mathbf{a}_{t-1}, \mathbf{a}_{t-2}), \\
    \mathbf{s}^\text{mot}_t &:= (\hat{\mathbf{g}}_t,\, \mathbf{v}_t,\, \boldsymbol{\omega}_t,\, \mathbf{q}_t,\, \dot{\mathbf{q}}_t,\, h_t,\ldots, \mathbf{p}^i_t, \mathbf{R}^i_t, \mathbf{v}^i_t, \boldsymbol{\omega}^i_t, \ldots ), \, \, i=1, \ldots, M \nonumber
\end{aligned}
\end{equation}
where the root orientation is represented by the projected gravity vector $\hat{\mathbf{g}}_t$, and $\mathbf{v}_t$ and $\boldsymbol{\omega}_t$ denote the root linear and angular velocities. $\mathbf{q}_t$ and $\dot{\mathbf{q}}_t$ are the joint positions and velocities, and $\mathbf{a}_{t-1}, \mathbf{a}_{t-2}$ denote the actions from previous timesteps. The motion state additionally includes the root height $h_t$. To ensure global pose invariance, the root linear and angular velocities, $\mathbf{v}_t$ and $\boldsymbol{\omega}_t$, are expressed in the root frame and the kinematic states ($\mathbf{p}^i_t, \mathbf{R}^i_t, \mathbf{v}^i_t, \boldsymbol{\omega}^i_t$) of the $M$ rigid bodies in a local path frame. The path frame is defined with the origin at the root, with the $x$-axis pointing forward and the $z$-axis upward.

To improve the sim-to-real transfer for Lima, we apply domain randomization to foot friction, mass properties, and external disturbance forces. Since these parameters are not observable on the physical robot, we adopt an asymmetric actor-critic architecture. The critic and forward map receive a privileged state $\mathbf{s}^\text{pri}_t$, which extends the actor state with the corresponding simulation parameters, enabling more accurate value estimation while keeping the actor restricted to observations available on hardware. For the SMPL humanoid, we do not apply domain randomization and instead employ a symmetric setup in which both the actor and the critic operate on the same actor state.

\paragraph{BFM Training} We train the BFM using FB-CPR~\cite{tirinzoni2025zero} within the parallel training framework of BFM-Zero~\cite{li2025bfm}, using 1024 simulation environments in parallel. To handle the resulting data throughput, we adopt the fast off-policy learning techniques of \cite{seoFastTD3SimpleFast2025} with an update-to-data (UTD) ratio of $1/64$. We use a 256-dimensional latent space and train the model for 3M gradient steps. Training the model takes $\sim$ three days on a single NVIDIA RTX~5090 GPU. Comprehensive implementation details, including hyperparameters, network architectures, and embodiment-specific settings, are provided in our supplemental material.  %

The adversarial regularization in FB-CPR is driven by a filtered version \cite{luo_perpetual_2023} of the AMASS dataset \cite{Mahmood2019AMASS}, with height corrections applied to account for discrepancies in the mocap data. For the Lima robot, the dataset is retargeted via an RL-based technique~\cite{muller2026reactor}. Following \citet{tirinzoni2025zero} and \citet{li2025bfm}, we address the imbalance in motion difficulty within AMASS by partitioning the dataset into \SI{10}{\second} chunks and re-weighting their sampling frequency. We periodically update these weights based on imitation accuracy, measured by the \emph{mean per-joint position error} (MPJPE). This ensures that challenging, highly dynamic sequences are prioritized by being sampled more frequently than simpler motions.
 
\paragraph{Latent Optimization} The latent optimization for motion tracking is performed over $24$ iterations using $N=128$ parallel environments. We use the Adam optimizer~\cite{kingmaAdamMethodStochastic2015} with a learning rate of $0.00625$ and modified momentum parameters $\beta=(0.8, 0.99)$. For all rollouts, we use the discount factor $\gamma=0.97$ and the exploration noise standard deviation $\sigma=0.0125$. On a single desktop equipped with an NVIDIA RTX 5090, our framework optimizes a full tracking trajectory in a few minutes.

\section{Results}

We evaluate the dense motion tracking performance on the test split introduced by \citet{tirinzoni2025zero}, which contains 990 motion sequences curated from the AMASS dataset \cite{Mahmood2019AMASS}. Together, these sequences comprise more than three hours of diverse human motion and provide a rigorous benchmark for both the SMPL humanoid and Lima robot, using retargeted motions for the latter. We apply the same data preprocessing pipeline used during training, as described in \secref{sec:implementation_details}.

For the evaluation of sparse motion tracking, we extract informative keyframes from the kinematic reference motions in the test set using a Difference-of-Gaussians (DoG) prominence heuristic, inspired by feature extraction in computer vision~\cite{lowe2004distinctive}. Specifically, we compute the sum of squared joint velocities and subtract a heavily smoothed version of this signal from a lightly smoothed one. The resulting prominence signal highlights salient motion events, such as rapid transitions and brief pauses. We then extract keyframes by selecting local maxima while enforcing a minimum temporal spacing $\Delta t$. By varying $\Delta t$, we can directly control the sparsity of the test set and achieve a desired median keyframe rate. Full details of the extraction procedure are provided in the accompanying supplemental material. 

\subsection{Metrics}
We evaluate tracking quality along three axes: motion preservation, tracking accuracy, and motion smoothness.

\paragraph{Motion Preservation} To assess whether the overall style and temporal structure of the reference motion are preserved, we use the Earth Mover's Distance (EMD)~\cite{emd2000} and Dynamic Time Warping (DTW)~\cite{sakoe1978dynamic}.
EMD measures the optimal transport cost between the set of poses visited during tracking and the reference poses, where each pose is represented by joint positions and 6D orientations in the local path frame (see~\secref{sec:implementation_details}).
DTW measures the $L_1$ distance on joint angles in radians under optimal temporal alignment, normalized by the warping path length.
Because EMD and DTW are alignment-tolerant, we evaluate them against the full dense reference even in the sparse keyframe setting.

\paragraph{Tracking Accuracy} We quantify tracking accuracy using two metrics based on the Euclidean distance between simulated and reference joint Cartesian positions in the local path frame. First, the \emph{mean per-joint position error} (MPJPE) averages this error across all motions, joints, and frames. Second, the \emph{mean max per-joint position error} (MMPJPE) captures the worst-case deviation by identifying the maximum error per joint within each motion sequence and averaging these peak errors across the entire dataset. Consequently, MMPJPE serves as a continuous, threshold-free proxy for tracking success rate~\cite{luo_perpetual_2023}. In the sparse setting, both metrics are evaluated only at keyframes.

\paragraph{Motion Smoothness} To quantify motion smoothness, we report \emph{mean per-joint acceleration error} (MPJAE), which averages the error between the simulated and reference joint linear acceleration in the local path frame.

\subsection{Dense Motion Tracking}
\begin{figure}
    \centering
    \includegraphics{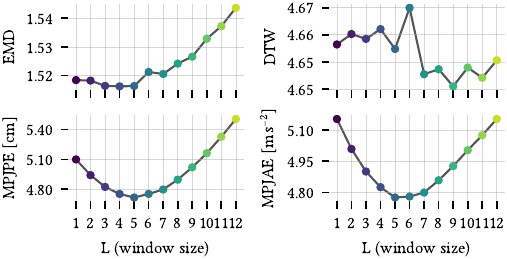}
    \caption{
        \textbf{Lookahead Horizon on ER Baseline.} Tracking performance peaks at a window size of $L=5$ and degrades for larger horizons, illustrating the coherence-precision trade-off discussed in \tabref{tab:dense}.
    }
    \label{fig:er-ablation}
\end{figure}

\begin{figure}
    \centering
    \includegraphics[width=\linewidth, trim={0 8pt 0 15pt}, clip]{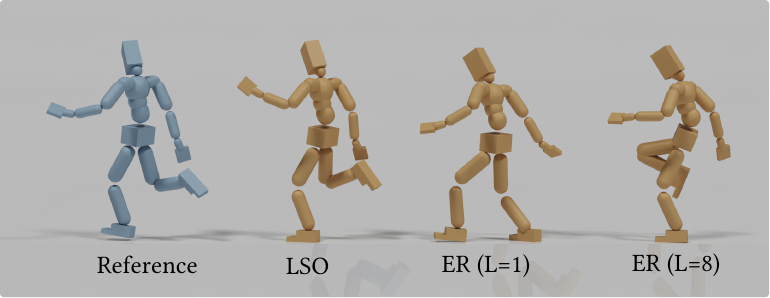}
    \caption{\textbf{Qualitative Comparison.} Comparison of \method against ER~(L=1) and ER~(L=8) for tracking a kicking reference motion. A short horizon lacks the necessary context to prepare for the kick, exhibiting a lag, whereas the long horizon averages too far ahead, leading to a premature kicking motion.}
    \label{fig:ER_qual_comparison}
\end{figure}

We compare our method against prior BFM-based baselines~\cite{tirinzoni2025zero,li2025bfm}, which approximate motion tracking by modeling the target sequence over a short horizon as a stationary behavior using the Empirical Reward (ER) introduced in~\secref{sec:fb}. This zero-shot ER formulation introduces the look-ahead horizon $L$ as a critical hyperparameter that requires proper tuning. To ensure a rigorous comparison, we ablate the horizon length in~\figref{fig:er-ablation} and visualize its effect on a dynamic kick sequence in~\figref{fig:ER_qual_comparison} and the supplemental video. A short horizon ($L=1$) lacks the anticipatory context needed to prepare for the kick, causing the executed motion to lag. Conversely, a long horizon ($L=8$) averages too far ahead, executing the kick prematurely. We find that $L=5$ strikes the optimal balance between anticipation and precision. Consequently, we establish the ER formulation with a horizon of $5$ as our strongest baseline for evaluating our proposed optimization framework.

With the optimal zero-shot baseline established, we evaluate the performance of our latent sequence optimization (\method) framework, reporting results in \tabref{tab:dense}. To assess the impact of temporal structure, we compare the baseline against three variants of our approach, controlled by the exploration noise parameter $\beta$. While the base variant ($\beta=0$) uses white noise to explore latents independently at each timestep, we inject temporally correlated pink ($\beta=1$) and red ($\beta=2$) noise to explicitly encourage smoother trajectories. Our independent latent optimization approach \method ($\beta=0$) substantially improves positional tracking accuracy over the ER baseline. However, optimizing independent variables at every timestep introduces slight high-frequency jitter, reflected in a higher acceleration error. By incorporating temporal structure into the optimization via correlated exploration noise, we successfully overcome this limitation. Pink noise ($\beta=1$) achieves the best overall tracking fidelity across all positional metrics. Furthermore, imposing stronger temporal correlations via red noise ($\beta=2$) yields the smoothest motion, achieving the lowest acceleration error, at the cost of marginally reduced tracking precision compared to $\beta=1$.

\begin{table}
    \footnotesize
    \caption{
        \textbf{Dense Motion Tracking Evaluation.} We compare \method against the ER baseline ($L=5$). While $\beta=0$ uses white noise, injecting temporal structure via pink ($\beta=1$) or red ($\beta=2$) noise allows \method to surpass the baseline. The best scores are bold, the second-best scores are underlined.
    }
    \label{tab:dense}
    \begin{tabular}{llrrrrr}
    \toprule
    && EMD $\downarrow$ & DTW $\downarrow$ & MPJPE $\downarrow$ & MMPJPE $\downarrow$ & MPJAE $\downarrow$ \\
    && & & [\SI{}{\centi\meter}] & [\SI{}{\centi\meter}] & [\SI{}{\meter\per\second\squared}] \\
    \midrule
    ER & ($L=5$)          & 1.52             & 4.66             & 4.7  & 40.6  & \underline{4.77}  \\
    \addlinespace
    \multirow{3}{*}{\makecell[l]{\method \\ (ours)}}
     & ($\beta=0$)  &           {1.29} & {4.07}           & 3.8  & 35.3  & 6.50  \\
     & ($\beta=1$)  & \textbf{1.17}    & \textbf{3.77}    & \textbf{3.4} & \textbf{32.6} & 5.32  \\
     & ($\beta=2$)& \underline{1.18} & \underline{3.78} & \underline{3.5}  & \underline{33.6}  & \textbf{4.64}  \\
    \bottomrule
    \end{tabular}
\end{table}

\subsection{Sparse Keyframe Tracking}
\label{sec:eval_sparse}

\paragraph{Baselines.}
Unlike the dense setting, where ER~(\secref{sec:fb}) provides an established zero-shot baseline, no equivalent exists for sparse keyframe tracking with BFMs. Because the sparse tracking objective is highly under-determined, imposing structural priors is essential to regularize the optimization and synthesize plausible in-between motion. We therefore ablate \method against alternative strategies for imposing this structure. The simplest baseline, which we term \emph{Piecewise Constant}, assigns a single latent to each keyframe and holds it constant throughout the timesteps between consecutive keyframes. To explicitly encourage smoother transitions, an alternative baseline, \emph{Spherical Spline}, parameterizes the latent sequence using continuous spherical splines. For both formulations, we initialize the keyframe latents using their zero-shot backward encodings $B(\mathbf{g}_{t_k})$ and optimize them using the same policy gradient (PG) algorithm described in~\secref{sec:latent_optimize}. We describe the detailed implementation of both baselines in our supplemental material.

\begin{figure}[htbp]
    \centering
    \includegraphics{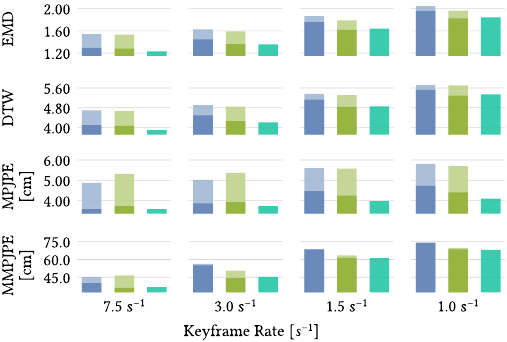}\\[0.3em]
    \begin{tikzpicture}
    \begin{axis}[
        hide axis,
        xmin=0, xmax=1, ymin=0, ymax=1,
        legend columns=3,
        legend style={
            draw=none, fill=none,
            font=\footnotesize,
            /tikz/every even column/.append style={column sep=6pt},
        },
    ]
    \addlegendimage{only marks, mark=square*, mark size=3pt, fill={rgb,1:red,0.621;green,0.704;blue,0.826}, draw=none}
    \addlegendentry{Piecewise Constant}
    \addlegendimage{only marks, mark=square*, mark size=3pt, fill={rgb,1:red,0.368;green,0.507;blue,0.710}, draw=none}
    \addlegendentry{Piecewise Constant + PG}
    \addlegendimage{empty legend}  %
    \addlegendentry{ }
    \addlegendimage{only marks, mark=square*, mark size=3pt, fill={rgb,1:red,0.736;green,0.815;blue,0.517}, draw=none}
    \addlegendentry{Spherical Spline}
    \addlegendimage{only marks, mark=square*, mark size=3pt, fill={rgb,1:red,0.560;green,0.692;blue,0.195}, draw=none}
    \addlegendentry{Spherical Spline + PG}
    \addlegendimage{only marks, mark=square*, mark size=3pt, fill={rgb,1:red,0.098;green,0.761;blue,0.631}, draw=none}
    \addlegendentry{LSO (ours)}
    \end{axis}
    \end{tikzpicture}
    \caption{\textbf{Sparse Keyframe Tracking.} Zero-shot initializations (lighter bars, behind) versus their optimized variants (darker bars, in front). Each column corresponds to a different keyframe density. \method with correlated noise achieves the best trade-off between tracking accuracy and motion quality. Lower is better.}\label{fig:sparse-bars}
\end{figure}

\begin{figure}
    \centering
    \includegraphics[width=\linewidth]{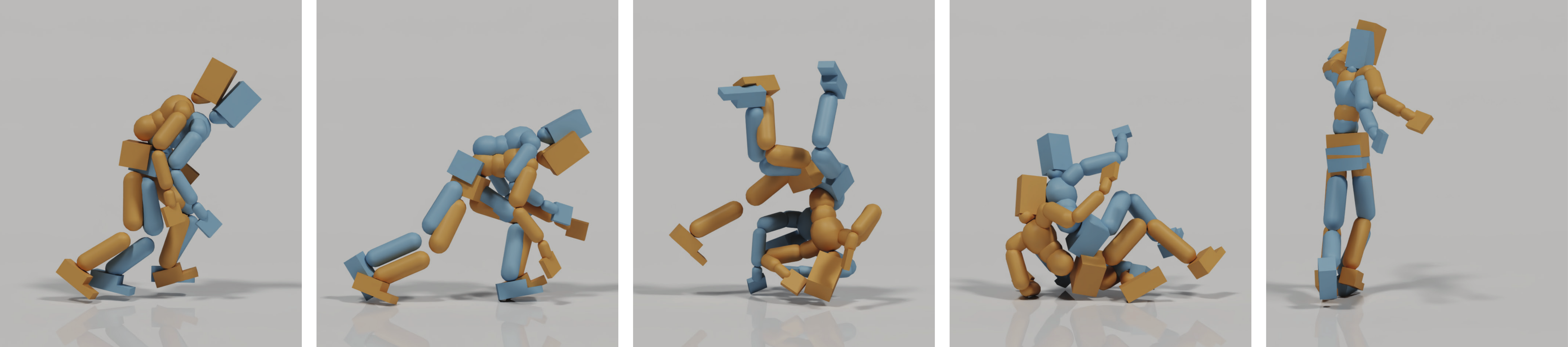}
    \caption{\textbf{Qualitative Sparse Motion Tracking.} \method accurately tracks a sparse keyframe sequence of an underlying, highly dynamic and contact-rich motion like a roll. See supplemental video for the full dynamic sequence. }
    \label{fig:sparse_tracking_roll}
\end{figure}

\paragraph{Evaluation.}
We compare \method against the two baselines, both with and without latent optimization, as shown in \figref{fig:sparse-bars} and the supplemental video. Across all methods, latent optimization significantly improves overall tracking performance and motion quality compared to the zero-shot initializations, underscoring the critical importance of the optimization step. Although these alternative formulations achieve competitive results, our approach outperforms them across most scenarios. Specifically, while the \emph{Piecewise Constant} formulation performs reasonably well for denser keyframes, its performance degrades in sparser settings, as optimization ultimately cannot overcome the abrupt latent switches inherent to its structure. Conversely, while the \emph{Spherical Spline} parameterization successfully smooths out these transitions, its constrained structure proves less effective for dense motions. In contrast, \method strikes a good balance between tracking accuracy and natural motion generation. We highlight this capability with a challenging rolling motion tracked from a sparse sequence of keyframes, visualized in \figref{fig:sparse_tracking_roll} and the supplemental video. Ultimately, \method proves to be a general framework that seamlessly extends our dense tracking success to the sparse domain, demonstrating the universal effectiveness of flexible trajectories guided by correlated noise.

\subsection{Robot Deployment}
We demonstrate qualitatively that our method can be transferred to a real robotic platform. See the supplemental video for results. The robot can execute dynamic and contact-rich motions such as running and dancing, and is able to react to perturbations and disturbances. Notably, the method is also able to get up from a fall: starting from an unplanned fallen state, the method is able to generate a contact-rich getting-up sequence that can be executed by the physical robot while staying true to the training motion distribution.

\section{Conclusion}

In this work, we bridge the gap between the generalist capabilities of BFMs and high-fidelity motion tracking of physics-based characters by moving from single latent selection to temporal latent optimization.

A key advantage of BFMs is their ability to produce motions that are grounded in physics and guided by a large-scale dataset, to interpolate sparse keyframes or recover from large perturbations. For instance, BFMs can produce physics-informed, plausible motion even if the character state deviates significantly from a tracking reference. This includes adding additional gait cycles between walking keyframes as required, or interpolating highly dynamic motions. 
This is particularly useful when transferring motions to a physical robot. Inevitable sim-to-real gaps mean that the nominal behavior is never perfectly matched. Even if the robot is perturbed, as we show in our supplemental video, the BFM is able to synthesize natural-looking, contact-rich getting-up sequences. 

\paragraph{Limitations.} Our proposed latent sequence optimization method is too computationally expensive for interactive animation workflows or real-time computation onboard a robot: we currently optimize the latent sequences offline prior to execution. This prevents its use for character or robot interactions with uncertain and complex environments, where the real-time adaptation of latent sequences to environmental constraints is a necessity. Moreover, BFMs trained with FB-CPR require a large representative motion dataset, which is not readily available for non-humanoid characters or for cartoon-like characters with caricatured motions.

\paragraph{Future Work.} 

BFMs have yet to be explored in depth within the research community, and while results to date show promising potential, there are many unexplored avenues. A natural extension of this work would be to allow for keyframes to define partial states, similar to recent approaches for tracking tasks with partial inputs \cite{tessler2024maskedmimic}. For example, only given end-effector positions as targets, BFMs could be leveraged to define the full robot pose. Another open question is how to support global (world-frame) tracking targets, which would be required for temporally extended tasks, e.g., loco-manipulation~\cite{luo2024omnigrasp} or overcoming challenging terrain~\cite{xu2025parkour}. Our proposed latent optimization has the potential to support such tasks.

\bibliographystyle{ACM-Reference-Format}
\bibliography{bibliography}

\begin{thebibliography}{59}


\ifx \showCODEN    \undefined \def \showCODEN     #1{\unskip}     \fi
\ifx \showISBNx    \undefined \def \showISBNx     #1{\unskip}     \fi
\ifx \showISBNxiii \undefined \def \showISBNxiii  #1{\unskip}     \fi
\ifx \showISSN     \undefined \def \showISSN      #1{\unskip}     \fi
\ifx \showLCCN     \undefined \def \showLCCN      #1{\unskip}     \fi
\ifx \shownote     \undefined \def \shownote      #1{#1}          \fi
\ifx \showarticletitle \undefined \def \showarticletitle #1{#1}   \fi
\ifx \showURL      \undefined \def \showURL       {\relax}        \fi
\providecommand\bibfield[2]{#2}
\providecommand\bibinfo[2]{#2}
\providecommand\natexlab[1]{#1}
\providecommand\showeprint[2][]{arXiv:#2}

\bibitem[Agrawal et~al\mbox{.}(2024)]%
        {agrawal2024skel}
\bibfield{author}{\bibinfo{person}{Dhruv Agrawal}, \bibinfo{person}{Jakob
  Buhmann}, \bibinfo{person}{Dominik Borer}, \bibinfo{person}{Robert~W Sumner},
  {and} \bibinfo{person}{Martin Guay}.} \bibinfo{year}{2024}\natexlab{}.
\newblock \showarticletitle{Skel-betweener: a neural motion rig for interactive
  motion authoring}.
\newblock \bibinfo{journal}{\emph{ACM Transactions on Graphics (TOG)}}
  \bibinfo{volume}{43}, \bibinfo{number}{6} (\bibinfo{year}{2024}),
  \bibinfo{pages}{1--11}.
\newblock


\bibitem[Bagatella et~al\mbox{.}(2026a)]%
        {bagatellaTDJEPALatentpredictiveRepresentations2025}
\bibfield{author}{\bibinfo{person}{Marco Bagatella}, \bibinfo{person}{Matteo
  Pirotta}, \bibinfo{person}{Ahmed Touati}, \bibinfo{person}{Alessandro
  Lazaric}, {and} \bibinfo{person}{Andrea Tirinzoni}.}
  \bibinfo{year}{2026}\natexlab{a}.
\newblock \showarticletitle{{TD}-{JEPA}: Latent-predictive Representations for
  Zero-Shot Reinforcement Learning}. In \bibinfo{booktitle}{\emph{The
  Fourteenth International Conference on Learning Representations}}.
\newblock
\urldef\tempurl%
\url{https://openreview.net/forum?id=SzXDuBN8M1}
\showURL{%
\tempurl}


\bibitem[Bagatella et~al\mbox{.}(2026b)]%
        {bagatella2026softforwardbackwardrepresentationszeroshot}
\bibfield{author}{\bibinfo{person}{Marco Bagatella}, \bibinfo{person}{Thomas
  Rupf}, \bibinfo{person}{Georg Martius}, {and} \bibinfo{person}{Andreas
  Krause}.} \bibinfo{year}{2026}\natexlab{b}.
\newblock \bibinfo{title}{Soft Forward-Backward Representations for Zero-shot
  Reinforcement Learning with General Utilities}.
\newblock
\showeprint[arxiv]{2602.06769}~[cs.LG]
\urldef\tempurl%
\url{https://arxiv.org/abs/2602.06769}
\showURL{%
\tempurl}


\bibitem[Cathomen et~al\mbox{.}(2025)]%
        {cathomenDivideDiscoverDeploy2025}
\bibfield{author}{\bibinfo{person}{Rafael Cathomen}, \bibinfo{person}{Mayank
  Mittal}, \bibinfo{person}{Marin Vlastelica}, {and} \bibinfo{person}{Marco
  Hutter}.} \bibinfo{year}{2025}\natexlab{}.
\newblock \showarticletitle{Divide, Discover, Deploy: Factorized Skill Learning
  with Symmetry and Style Priors}. In \bibinfo{booktitle}{\emph{Conference on
  Robot Learning}}.
\newblock


\bibitem[Eberhard et~al\mbox{.}(2023)]%
        {eberhard2023pink}
\bibfield{author}{\bibinfo{person}{Onno Eberhard}, \bibinfo{person}{Jakob
  Hollenstein}, \bibinfo{person}{Cristina Pinneri}, {and}
  \bibinfo{person}{Georg Martius}.} \bibinfo{year}{2023}\natexlab{}.
\newblock \showarticletitle{Pink noise is all you need: Colored noise
  exploration in deep reinforcement learning}. In \bibinfo{booktitle}{\emph{The
  Eleventh International Conference on Learning Representations}}.
\newblock


\bibitem[Eysenbach et~al\mbox{.}(2021)]%
        {eysenbachInformationGeometryUnsupervised2021}
\bibfield{author}{\bibinfo{person}{Benjamin Eysenbach}, \bibinfo{person}{Ruslan
  Salakhutdinov}, {and} \bibinfo{person}{Sergey Levine}.}
  \bibinfo{year}{2021}\natexlab{}.
\newblock \bibinfo{title}{The {{Information Geometry}} of {{Unsupervised
  Reinforcement Learning}}}.
\newblock
\showeprint[arxiv]{2110.02719}~[cs]
\href{https://doi.org/10.48550/arXiv.2110.02719}{doi:\nolinkurl{10.48550/arXiv.2110.02719}}


\bibitem[Fussell et~al\mbox{.}(2021)]%
        {fussell2021supertrack}
\bibfield{author}{\bibinfo{person}{Levi Fussell}, \bibinfo{person}{Kevin
  Bergamin}, {and} \bibinfo{person}{Daniel Holden}.}
  \bibinfo{year}{2021}\natexlab{}.
\newblock \showarticletitle{Supertrack: Motion tracking for physically
  simulated characters using supervised learning}.
\newblock \bibinfo{journal}{\emph{ACM Transactions on Graphics (TOG)}}
  \bibinfo{volume}{40}, \bibinfo{number}{6} (\bibinfo{year}{2021}),
  \bibinfo{pages}{1--13}.
\newblock


\bibitem[Grandia et~al\mbox{.}(2024)]%
        {grandia2024}
\bibfield{author}{\bibinfo{person}{Ruben Grandia}, \bibinfo{person}{Espen
  Knoop}, \bibinfo{person}{Michael~A. Hopkins}, \bibinfo{person}{Georg
  Wiedebach}, \bibinfo{person}{Jared Bishop}, \bibinfo{person}{Steven Pickles},
  \bibinfo{person}{David Müller}, {and} \bibinfo{person}{Moritz Bächer}.}
  \bibinfo{year}{2024}\natexlab{}.
\newblock \showarticletitle{{Design and Control of a Bipedal Robotic
  Character}}. In \bibinfo{booktitle}{\emph{Proceedings of Robotics: Science
  and Systems}}. \bibinfo{address}{Delft, Netherlands}.
\newblock
\href{https://doi.org/10.15607/RSS.2024.XX.103}{doi:\nolinkurl{10.15607/RSS.2024.XX.103}}


\bibitem[Harvey et~al\mbox{.}(2020)]%
        {harvey2020robust}
\bibfield{author}{\bibinfo{person}{F{\'e}lix~G Harvey}, \bibinfo{person}{Mike
  Yurick}, \bibinfo{person}{Derek Nowrouzezahrai}, {and}
  \bibinfo{person}{Christopher Pal}.} \bibinfo{year}{2020}\natexlab{}.
\newblock \showarticletitle{Robust motion in-betweening}.
\newblock \bibinfo{journal}{\emph{ACM Transactions on Graphics (TOG)}}
  \bibinfo{volume}{39}, \bibinfo{number}{4} (\bibinfo{year}{2020}),
  \bibinfo{pages}{60--1}.
\newblock


\bibitem[Huang et~al\mbox{.}(2025c)]%
        {huang2025adaptive}
\bibfield{author}{\bibinfo{person}{Tao Huang}, \bibinfo{person}{Huayi Wang},
  \bibinfo{person}{Junli Ren}, \bibinfo{person}{Kangning Yin},
  \bibinfo{person}{Zirui Wang}, \bibinfo{person}{Xiao Chen},
  \bibinfo{person}{Feiyu Jia}, \bibinfo{person}{Wentao Zhang},
  \bibinfo{person}{Jungfeng Long}, \bibinfo{person}{Jingbo Wang}, {and}
  \bibinfo{person}{Jiangmiao Pang}.} \bibinfo{year}{2025}\natexlab{c}.
\newblock \showarticletitle{Towards Adaptable Humanoid Control via Adaptive
  Motion Tracking}.
\newblock \bibinfo{journal}{\emph{arXiv preprint arXiv:2510.14454}}
  (\bibinfo{year}{2025}).
\newblock


\bibitem[Huang et~al\mbox{.}(2025b)]%
        {huang2025diffusecloc}
\bibfield{author}{\bibinfo{person}{Xiaoyu Huang}, \bibinfo{person}{Takara
  Truong}, \bibinfo{person}{Yunbo Zhang}, \bibinfo{person}{Fangzhou Yu},
  \bibinfo{person}{Jean~Pierre Sleiman}, \bibinfo{person}{Jessica Hodgins},
  \bibinfo{person}{Koushil Sreenath}, {and} \bibinfo{person}{Farbod
  Farshidian}.} \bibinfo{year}{2025}\natexlab{b}.
\newblock \showarticletitle{Diffuse-CLoC: Guided Diffusion for Physics-based
  Character Look-ahead Control}.
\newblock \bibinfo{journal}{\emph{ACM Transactions on Graphics (TOG)}}
  \bibinfo{volume}{44}, \bibinfo{number}{4}, Article \bibinfo{articleno}{132}
  (\bibinfo{date}{July} \bibinfo{year}{2025}), \bibinfo{numpages}{12}~pages.
\newblock
\showISSN{0730-0301}
\href{https://doi.org/10.1145/3731206}{doi:\nolinkurl{10.1145/3731206}}


\bibitem[Huang et~al\mbox{.}(2025a)]%
        {huang2025modskill}
\bibfield{author}{\bibinfo{person}{Yiming Huang}, \bibinfo{person}{Zhiyang
  Dou}, {and} \bibinfo{person}{Lingjie Liu}.} \bibinfo{year}{2025}\natexlab{a}.
\newblock \showarticletitle{Modskill: Physical character skill modularization}.
  In \bibinfo{booktitle}{\emph{IEEE/CVF International Conference on Computer
  Vision}}.
\newblock


\bibitem[Hugessen et~al\mbox{.}(2025)]%
        {hugessenZeroShotConstraintSatisfaction2025}
\bibfield{author}{\bibinfo{person}{Adriana Hugessen}, \bibinfo{person}{Harley
  Wiltzer}, \bibinfo{person}{Cyrus Neary}, \bibinfo{person}{Amy Zhang}, {and}
  \bibinfo{person}{Glen Berseth}.} \bibinfo{year}{2025}\natexlab{}.
\newblock \showarticletitle{Zero-{{Shot Constraint Satisfaction}} with
  {{Forward- Backward Representations}}}. In \bibinfo{booktitle}{\emph{Workshop
  on {{Reinforcement Learning Beyond Rewards}} @ {{Reinforcement Learning
  Conference}} 2025}}.
\newblock


\bibitem[Jajoo et~al\mbox{.}(2025)]%
        {jajooRegularizedLatentDynamics2025}
\bibfield{author}{\bibinfo{person}{Pranaya Jajoo}, \bibinfo{person}{Harshit
  Sikchi}, \bibinfo{person}{Siddhant Agarwal}, \bibinfo{person}{Amy Zhang},
  \bibinfo{person}{Scott Niekum}, {and} \bibinfo{person}{Martha White}.}
  \bibinfo{year}{2025}\natexlab{}.
\newblock \showarticletitle{Regularized {{Latent Dynamics Prediction}} Is a
  {{Strong Baseline For Behavioral Foundation Models}}}. In
  \bibinfo{booktitle}{\emph{Workshop on {{Reinforcement Learning Beyond
  Rewards}} @ {{Reinforcement Learning Conference}} 2025}}.
\newblock


\bibitem[Kingma and Ba(2015)]%
        {kingmaAdamMethodStochastic2015}
\bibfield{author}{\bibinfo{person}{Diederik~P. Kingma} {and}
  \bibinfo{person}{Jimmy Ba}.} \bibinfo{year}{2015}\natexlab{}.
\newblock \showarticletitle{Adam: A Method for Stochastic Optimization}. In
  \bibinfo{booktitle}{\emph{3rd International Conference on Learning
  Representations, {ICLR} 2015, San Diego, CA, USA, May 7-9, 2015, Conference
  Track Proceedings}}.
\newblock
\urldef\tempurl%
\url{http://arxiv.org/abs/1412.6980}
\showURL{%
\tempurl}


\bibitem[Kool et~al\mbox{.}(2019)]%
        {kool2019buy}
\bibfield{author}{\bibinfo{person}{Wouter Kool}, \bibinfo{person}{Herke van
  Hoof}, {and} \bibinfo{person}{Max Welling}.} \bibinfo{year}{2019}\natexlab{}.
\newblock \showarticletitle{Buy 4 reinforce samples, get a baseline for free!}
\newblock \bibinfo{journal}{\emph{Deep Reinforcement Learning Meets Structured
  Prediction Workshop at ICLR}} (\bibinfo{year}{2019}).
\newblock


\bibitem[Li et~al\mbox{.}(2025)]%
        {li2025robotic}
\bibfield{author}{\bibinfo{person}{Chenhao Li}, \bibinfo{person}{Andreas
  Krause}, {and} \bibinfo{person}{Marco Hutter}.}
  \bibinfo{year}{2025}\natexlab{}.
\newblock \showarticletitle{Robotic world model: A neural network simulator for
  robust policy optimization in robotics}.
\newblock \bibinfo{journal}{\emph{arXiv preprint arXiv:2501.10100}}
  (\bibinfo{year}{2025}).
\newblock


\bibitem[Li et~al\mbox{.}(2026)]%
        {li2025bfm}
\bibfield{author}{\bibinfo{person}{Yitang Li}, \bibinfo{person}{Zhengyi Luo},
  \bibinfo{person}{Tonghe Zhang}, \bibinfo{person}{Cunxi Dai},
  \bibinfo{person}{Anssi Kanervisto}, \bibinfo{person}{Andrea Tirinzoni},
  \bibinfo{person}{Haoyang Weng}, \bibinfo{person}{Kris Kitani},
  \bibinfo{person}{Mateusz Guzek}, \bibinfo{person}{Ahmed Touati},
  \bibinfo{person}{Alessandro Lazaric}, \bibinfo{person}{Matteo Pirotta}, {and}
  \bibinfo{person}{Guanya Shi}.} \bibinfo{year}{2026}\natexlab{}.
\newblock \showarticletitle{{BFM}-Zero: A Promptable Behavioral Foundation
  Model for Humanoid Control Using Unsupervised Reinforcement Learning}. In
  \bibinfo{booktitle}{\emph{The Fourteenth International Conference on Learning
  Representations}}.
\newblock
\urldef\tempurl%
\url{https://openreview.net/forum?id=jkhl2oI0g5}
\showURL{%
\tempurl}


\bibitem[Liao et~al\mbox{.}(2025)]%
        {liao2025beyondmimic}
\bibfield{author}{\bibinfo{person}{Qiayuan Liao}, \bibinfo{person}{Takara~E
  Truong}, \bibinfo{person}{Xiaoyu Huang}, \bibinfo{person}{Guy Tevet},
  \bibinfo{person}{Koushil Sreenath}, {and} \bibinfo{person}{C~Karen Liu}.}
  \bibinfo{year}{2025}\natexlab{}.
\newblock \showarticletitle{Beyondmimic: From motion tracking to versatile
  humanoid control via guided diffusion}.
\newblock \bibinfo{journal}{\emph{arXiv preprint arXiv:2508.08241}}
  (\bibinfo{year}{2025}).
\newblock


\bibitem[Loper et~al\mbox{.}(2015)]%
        {loper2015smpl}
\bibfield{author}{\bibinfo{person}{Matthew Loper}, \bibinfo{person}{Naureen
  Mahmood}, \bibinfo{person}{Javier Romero}, \bibinfo{person}{Gerard
  Pons-Moll}, {and} \bibinfo{person}{Michael~J. Black}.}
  \bibinfo{year}{2015}\natexlab{}.
\newblock \showarticletitle{SMPL: a skinned multi-person linear model}.
\newblock \bibinfo{journal}{\emph{ACM Transactions on Graphics (TOG)}}
  \bibinfo{volume}{34}, \bibinfo{number}{6}, Article \bibinfo{articleno}{248}
  (\bibinfo{date}{Nov.} \bibinfo{year}{2015}), \bibinfo{numpages}{16}~pages.
\newblock
\showISSN{0730-0301}
\href{https://doi.org/10.1145/2816795.2818013}{doi:\nolinkurl{10.1145/2816795.2818013}}


\bibitem[Lowe(2004)]%
        {lowe2004distinctive}
\bibfield{author}{\bibinfo{person}{David~G Lowe}.}
  \bibinfo{year}{2004}\natexlab{}.
\newblock \showarticletitle{Distinctive image features from scale-invariant
  keypoints}.
\newblock \bibinfo{journal}{\emph{International journal of computer vision}}
  \bibinfo{volume}{60}, \bibinfo{number}{2} (\bibinfo{year}{2004}),
  \bibinfo{pages}{91--110}.
\newblock


\bibitem[Luo et~al\mbox{.}(2024)]%
        {luo2024omnigrasp}
\bibfield{author}{\bibinfo{person}{Zhengyi Luo}, \bibinfo{person}{Jinkun Cao},
  \bibinfo{person}{Sammy Christen}, \bibinfo{person}{Alexander Winkler},
  \bibinfo{person}{Kris Kitani}, {and} \bibinfo{person}{Weipeng Xu}.}
  \bibinfo{year}{2024}\natexlab{}.
\newblock \showarticletitle{Omnigrasp: Grasping diverse objects with simulated
  humanoids}.
\newblock \bibinfo{journal}{\emph{Advances in Neural Information Processing
  Systems}}  \bibinfo{volume}{37} (\bibinfo{year}{2024}),
  \bibinfo{pages}{2161--2184}.
\newblock


\bibitem[Luo et~al\mbox{.}(2023)]%
        {luo_perpetual_2023}
\bibfield{author}{\bibinfo{person}{Zhengyi Luo}, \bibinfo{person}{Jinkun Cao},
  \bibinfo{person}{Alexander Winkler}, \bibinfo{person}{Kris Kitani}, {and}
  \bibinfo{person}{Weipeng Xu}.} \bibinfo{year}{2023}\natexlab{}.
\newblock \showarticletitle{Perpetual {Humanoid} {Control} for {Real}-time
  {Simulated} {Avatars}}. In \bibinfo{booktitle}{\emph{2023 {IEEE}/{CVF}
  {International} {Conference} on {Computer} {Vision}}}.
\newblock
\href{https://doi.org/10.1109/ICCV51070.2023.01000}{doi:\nolinkurl{10.1109/ICCV51070.2023.01000}}


\bibitem[Luo et~al\mbox{.}(2025)]%
        {luo2025sonic}
\bibfield{author}{\bibinfo{person}{Zhengyi Luo}, \bibinfo{person}{Ye Yuan},
  \bibinfo{person}{Tingwu Wang}, \bibinfo{person}{Chenran Li},
  \bibinfo{person}{Sirui Chen}, \bibinfo{person}{Fernando Casta\~neda},
  \bibinfo{person}{Zi-Ang Cao}, \bibinfo{person}{Jiefeng Li},
  \bibinfo{person}{David Minor}, \bibinfo{person}{Qingwei Ben},
  \bibinfo{person}{Xingye Da}, \bibinfo{person}{Runyu Ding},
  \bibinfo{person}{Cyrus Hogg}, \bibinfo{person}{Lina Song},
  \bibinfo{person}{Edy Lim}, \bibinfo{person}{Eugene Jeong},
  \bibinfo{person}{Tairan He}, \bibinfo{person}{Haoru Xue},
  \bibinfo{person}{Wenli Xiao}, \bibinfo{person}{Zi Wang},
  \bibinfo{person}{Simon Yuen}, \bibinfo{person}{Jan Kautz},
  \bibinfo{person}{Yan Chang}, \bibinfo{person}{Umar Iqbal},
  \bibinfo{person}{Linxi Fan}, {and} \bibinfo{person}{Yuke Zhu}.}
  \bibinfo{year}{2025}\natexlab{}.
\newblock \showarticletitle{SONIC: Supersizing Motion Tracking for Natural
  Humanoid Whole-Body Control}.
\newblock \bibinfo{journal}{\emph{arXiv preprint arXiv:2511.07820}}
  (\bibinfo{year}{2025}).
\newblock


\bibitem[Mahmood et~al\mbox{.}(2019)]%
        {Mahmood2019AMASS}
\bibfield{author}{\bibinfo{person}{Naureen Mahmood}, \bibinfo{person}{Nima
  Ghorbani}, \bibinfo{person}{Nikolaus~F. Troje}, \bibinfo{person}{Gerard
  Pons-Moll}, {and} \bibinfo{person}{Michael~J. Black}.}
  \bibinfo{year}{2019}\natexlab{}.
\newblock \showarticletitle{{AMASS}: Archive of Motion Capture as Surface
  Shapes}. In \bibinfo{booktitle}{\emph{International Conference on Computer
  Vision}}. \bibinfo{pages}{5442--5451}.
\newblock


\bibitem[Makoviychuk et~al\mbox{.}(2021)]%
        {makoviychuk2021isaac}
\bibfield{author}{\bibinfo{person}{Viktor Makoviychuk}, \bibinfo{person}{Lukasz
  Wawrzyniak}, \bibinfo{person}{Yunrong Guo}, \bibinfo{person}{Michelle Lu},
  \bibinfo{person}{Kier Storey}, \bibinfo{person}{Miles Macklin},
  \bibinfo{person}{David Hoeller}, \bibinfo{person}{Nikita Rudin},
  \bibinfo{person}{Arthur Allshire}, \bibinfo{person}{Ankur Handa}, {and}
  \bibinfo{person}{Gavriel State}.} \bibinfo{year}{2021}\natexlab{}.
\newblock \showarticletitle{Isaac Gym: High Performance GPU-Based Physics
  Simulation For Robot Learning}. In \bibinfo{booktitle}{\emph{Thirty-fifth
  Conference on Neural Information Processing Systems Datasets and Benchmarks
  Track}}.
\newblock
\urldef\tempurl%
\url{https://openreview.net/forum?id=fgFBtYgJQX_}
\showURL{%
\tempurl}


\bibitem[M{\"u}ller et~al\mbox{.}(2026)]%
        {muller2026reactor}
\bibfield{author}{\bibinfo{person}{David M{\"u}ller}, \bibinfo{person}{Agon
  Serifi}, \bibinfo{person}{Sammy Christen}, \bibinfo{person}{Ruben Grandia},
  \bibinfo{person}{Espen Knoop}, {and} \bibinfo{person}{Moritz B{\"a}cher}.}
  \bibinfo{year}{2026}\natexlab{}.
\newblock \showarticletitle{ReActor: Reinforcement Learning for Physics-Aware
  Motion Retargeting}.
\newblock \bibinfo{journal}{\emph{arXiv preprint arXiv:2605.06593}}
  (\bibinfo{year}{2026}).
\newblock


\bibitem[Paraschos et~al\mbox{.}(2013)]%
        {paraschos2013probabilistic}
\bibfield{author}{\bibinfo{person}{Alexandros Paraschos},
  \bibinfo{person}{Christian Daniel}, \bibinfo{person}{Jan~R Peters}, {and}
  \bibinfo{person}{Gerhard Neumann}.} \bibinfo{year}{2013}\natexlab{}.
\newblock \showarticletitle{Probabilistic movement primitives}.
\newblock \bibinfo{journal}{\emph{Advances in neural information processing
  systems}}  \bibinfo{volume}{26} (\bibinfo{year}{2013}).
\newblock


\bibitem[Park et~al\mbox{.}(2024)]%
        {parkFoundationPoliciesHilbert2024}
\bibfield{author}{\bibinfo{person}{Seohong Park}, \bibinfo{person}{Tobias
  Kreiman}, {and} \bibinfo{person}{Sergey Levine}.}
  \bibinfo{year}{2024}\natexlab{}.
\newblock \showarticletitle{Foundation policies with Hilbert representations}.
  In \bibinfo{booktitle}{\emph{Proceedings of the 41st International Conference
  on Machine Learning}} (Vienna, Austria) \emph{(\bibinfo{series}{ICML'24})}.
  \bibinfo{publisher}{JMLR.org}, Article \bibinfo{articleno}{1609},
  \bibinfo{numpages}{25}~pages.
\newblock


\bibitem[Peng et~al\mbox{.}(2018)]%
        {pengDeepMimicExampleGuidedDeep2018}
\bibfield{author}{\bibinfo{person}{Xue~Bin Peng}, \bibinfo{person}{Pieter
  Abbeel}, \bibinfo{person}{Sergey Levine}, {and} \bibinfo{person}{Michiel
  van~de Panne}.} \bibinfo{year}{2018}\natexlab{}.
\newblock \showarticletitle{{{DeepMimic}}: {{Example-Guided Deep Reinforcement
  Learning}} of {{Physics-Based Character Skills}}}.
\newblock \bibinfo{journal}{\emph{ACM Transactions on Graphics (TOG)}}
  \bibinfo{volume}{37}, \bibinfo{number}{4} (\bibinfo{date}{Aug.}
  \bibinfo{year}{2018}), \bibinfo{pages}{1--14}.
\newblock
\showISSN{0730-0301, 1557-7368}
\showeprint[arxiv]{1804.02717}~[cs]
\href{https://doi.org/10.1145/3197517.3201311}{doi:\nolinkurl{10.1145/3197517.3201311}}


\bibitem[Peng et~al\mbox{.}(2022)]%
        {pengASELargeScaleReusable2022}
\bibfield{author}{\bibinfo{person}{Xue~Bin Peng}, \bibinfo{person}{Yunrong
  Guo}, \bibinfo{person}{Lina Halper}, \bibinfo{person}{Sergey Levine}, {and}
  \bibinfo{person}{Sanja Fidler}.} \bibinfo{year}{2022}\natexlab{}.
\newblock \showarticletitle{{{ASE}}: {{Large-Scale Reusable Adversarial Skill
  Embeddings}} for {{Physically Simulated Characters}}}.
\newblock \bibinfo{journal}{\emph{ACM Transactions on Graphics (TOG)}}
  \bibinfo{volume}{41}, \bibinfo{number}{4} (\bibinfo{date}{July}
  \bibinfo{year}{2022}), \bibinfo{pages}{1--17}.
\newblock
\showISSN{0730-0301, 1557-7368}
\showeprint[arxiv]{2205.01906}~[cs]
\href{https://doi.org/10.1145/3528223.3530110}{doi:\nolinkurl{10.1145/3528223.3530110}}


\bibitem[Peng et~al\mbox{.}(2021)]%
        {pengAMPAdversarialMotion2021}
\bibfield{author}{\bibinfo{person}{Xue~Bin Peng}, \bibinfo{person}{Ze Ma},
  \bibinfo{person}{Pieter Abbeel}, \bibinfo{person}{Sergey Levine}, {and}
  \bibinfo{person}{Angjoo Kanazawa}.} \bibinfo{year}{2021}\natexlab{}.
\newblock \showarticletitle{{{AMP}}: {{Adversarial Motion Priors}} for
  {{Stylized Physics-Based Character Control}}}.
\newblock \bibinfo{journal}{\emph{ACM Transactions on Graphics (TOG)}}
  \bibinfo{volume}{40}, \bibinfo{number}{4} (\bibinfo{date}{Aug.}
  \bibinfo{year}{2021}), \bibinfo{pages}{1--20}.
\newblock
\showISSN{0730-0301, 1557-7368}
\showeprint{2104.02180}
\href{https://doi.org/10.1145/3450626.3459670}{doi:\nolinkurl{10.1145/3450626.3459670}}


\bibitem[Pinneri et~al\mbox{.}(2021)]%
        {pinneriSampleefficientCrossEntropyMethod2020}
\bibfield{author}{\bibinfo{person}{Cristina Pinneri},
  \bibinfo{person}{Shambhuraj Sawant}, \bibinfo{person}{Sebastian Blaes},
  \bibinfo{person}{Jan Achterhold}, \bibinfo{person}{Joerg Stueckler},
  \bibinfo{person}{Michal Rolinek}, {and} \bibinfo{person}{Georg Martius}.}
  \bibinfo{year}{2021}\natexlab{}.
\newblock \showarticletitle{Sample-efficient cross-entropy method for real-time
  planning}. In \bibinfo{booktitle}{\emph{Conference on Robot Learning}}.
  \bibinfo{pages}{1049--1065}.
\newblock


\bibitem[Pirotta et~al\mbox{.}(2023)]%
        {pirottaFastImitationBehavior2023}
\bibfield{author}{\bibinfo{person}{Matteo Pirotta}, \bibinfo{person}{Andrea
  Tirinzoni}, \bibinfo{person}{Ahmed Touati}, \bibinfo{person}{Alessandro
  Lazaric}, {and} \bibinfo{person}{Yann Ollivier}.}
  \bibinfo{year}{2023}\natexlab{}.
\newblock \showarticletitle{Fast {{Imitation}} via {{Behavior Foundation
  Models}}}. In \bibinfo{booktitle}{\emph{The {{Twelfth International
  Conference}} on {{Learning Representations}}}}.
\newblock


\bibitem[Rubner et~al\mbox{.}(2000)]%
        {emd2000}
\bibfield{author}{\bibinfo{person}{Yossi Rubner}, \bibinfo{person}{Carlo
  Tomasi}, {and} \bibinfo{person}{Leonidas Guibas}.}
  \bibinfo{year}{2000}\natexlab{}.
\newblock \showarticletitle{The Earth Mover's Distance as a Metric for Image
  Retrieval}.
\newblock \bibinfo{journal}{\emph{International Journal of Computer Vision}}
  \bibinfo{volume}{40} (\bibinfo{date}{11} \bibinfo{year}{2000}),
  \bibinfo{pages}{99--121}.
\newblock
\href{https://doi.org/10.1023/A:1026543900054}{doi:\nolinkurl{10.1023/A:1026543900054}}


\bibitem[Rupf et~al\mbox{.}(2026)]%
        {rupfOptimisticTaskInference2025}
\bibfield{author}{\bibinfo{person}{Thomas Rupf}, \bibinfo{person}{Marco
  Bagatella}, \bibinfo{person}{Marin Vlastelica}, {and}
  \bibinfo{person}{Andreas Krause}.} \bibinfo{year}{2026}\natexlab{}.
\newblock \showarticletitle{Optimistic Task Inference for Behavior Foundation
  Models}. In \bibinfo{booktitle}{\emph{The Fourteenth International Conference
  on Learning Representations}}.
\newblock
\urldef\tempurl%
\url{https://openreview.net/forum?id=m5byThUSNE}
\showURL{%
\tempurl}


\bibitem[Sakoe and Chiba(1978)]%
        {sakoe1978dynamic}
\bibfield{author}{\bibinfo{person}{Hiroaki Sakoe} {and} \bibinfo{person}{Seibi
  Chiba}.} \bibinfo{year}{1978}\natexlab{}.
\newblock \showarticletitle{Dynamic programming algorithm optimization for
  spoken word recognition}.
\newblock \bibinfo{journal}{\emph{IEEE transactions on acoustics, speech, and
  signal processing}} \bibinfo{volume}{26}, \bibinfo{number}{1}
  (\bibinfo{year}{1978}), \bibinfo{pages}{43--49}.
\newblock


\bibitem[Seo et~al\mbox{.}(2025)]%
        {seoFastTD3SimpleFast2025}
\bibfield{author}{\bibinfo{person}{Younggyo Seo}, \bibinfo{person}{Carmelo
  Sferrazza}, \bibinfo{person}{Haoran Geng}, \bibinfo{person}{Michal Nauman},
  \bibinfo{person}{Zhao-Heng Yin}, {and} \bibinfo{person}{Pieter Abbeel}.}
  \bibinfo{year}{2025}\natexlab{}.
\newblock \bibinfo{title}{{{FastTD3}}: {{Simple}}, {{Fast}}, and {{Capable
  Reinforcement Learning}} for {{Humanoid Control}}}.
\newblock
\showeprint[arxiv]{2505.22642}~[cs]
\href{https://doi.org/10.48550/arXiv.2505.22642}{doi:\nolinkurl{10.48550/arXiv.2505.22642}}


\bibitem[Serifi et~al\mbox{.}(2024)]%
        {serifi2024vmp}
\bibfield{author}{\bibinfo{person}{Agon Serifi}, \bibinfo{person}{Ruben
  Grandia}, \bibinfo{person}{Espen Knoop}, \bibinfo{person}{Markus Gross},
  {and} \bibinfo{person}{Moritz B{\"a}cher}.} \bibinfo{year}{2024}\natexlab{}.
\newblock \showarticletitle{VMP: Versatile Motion Priors for Robustly Tracking
  Motion on Physical Characters}. In \bibinfo{booktitle}{\emph{Proceedings of
  the ACM SIGGRAPH/Eurographics Symposium on Computer Animation}} (Montreal,
  Quebec, Canada) \emph{(\bibinfo{series}{SCA '24})}.
  \bibinfo{publisher}{Eurographics Association}, \bibinfo{pages}{1–11}.
\newblock
\showISSN{1467-8659}
\href{https://doi.org/10.1111/cgf.15175}{doi:\nolinkurl{10.1111/cgf.15175}}


\bibitem[Shoemake(1985)]%
        {shoemake1985animating}
\bibfield{author}{\bibinfo{person}{Ken Shoemake}.}
  \bibinfo{year}{1985}\natexlab{}.
\newblock \showarticletitle{Animating rotation with quaternion curves}. In
  \bibinfo{booktitle}{\emph{Proceedings of the 12th annual conference on
  Computer graphics and interactive techniques}}. \bibinfo{pages}{245--254}.
\newblock


\bibitem[Sikchi et~al\mbox{.}(2025)]%
        {sikchiFastAdaptationBehavioral2025}
\bibfield{author}{\bibinfo{person}{Harshit Sikchi}, \bibinfo{person}{Andrea
  Tirinzoni}, \bibinfo{person}{Ahmed Touati}, \bibinfo{person}{Yingchen Xu},
  \bibinfo{person}{Anssi Kanervisto}, \bibinfo{person}{Scott Niekum},
  \bibinfo{person}{Amy Zhang}, \bibinfo{person}{Alessandro Lazaric}, {and}
  \bibinfo{person}{Matteo Pirotta}.} \bibinfo{year}{2025}\natexlab{}.
\newblock \showarticletitle{Fast Adaptation with Behavioral Foundation Models}.
  In \bibinfo{booktitle}{\emph{Reinforcement Learning Conference}}.
\newblock
\urldef\tempurl%
\url{https://openreview.net/forum?id=soeW8RGo1N}
\showURL{%
\tempurl}


\bibitem[Tessler et~al\mbox{.}(2024)]%
        {tessler2024maskedmimic}
\bibfield{author}{\bibinfo{person}{Chen Tessler}, \bibinfo{person}{Yunrong
  Guo}, \bibinfo{person}{Ofir Nabati}, \bibinfo{person}{Gal Chechik}, {and}
  \bibinfo{person}{Xue~Bin Peng}.} \bibinfo{year}{2024}\natexlab{}.
\newblock \showarticletitle{Maskedmimic: Unified physics-based character
  control through masked motion inpainting}.
\newblock \bibinfo{journal}{\emph{ACM Transactions on Graphics (TOG)}}
  \bibinfo{volume}{43}, \bibinfo{number}{6} (\bibinfo{year}{2024}),
  \bibinfo{pages}{1--21}.
\newblock


\bibitem[Tessler et~al\mbox{.}(2023)]%
        {tesslerCALMConditionalAdversarial2023}
\bibfield{author}{\bibinfo{person}{Chen Tessler}, \bibinfo{person}{Yoni
  Kasten}, \bibinfo{person}{Yunrong Guo}, \bibinfo{person}{Shie Mannor},
  \bibinfo{person}{Gal Chechik}, {and} \bibinfo{person}{Xue~Bin Peng}.}
  \bibinfo{year}{2023}\natexlab{}.
\newblock \showarticletitle{{{CALM}}: {{Conditional Adversarial Latent Models}}
  for {{Directable Virtual Characters}}}. In \bibinfo{booktitle}{\emph{Special
  {{Interest Group}} on {{Computer Graphics}} and {{Interactive Techniques
  Conference Conference Proceedings}}}}. \bibinfo{pages}{1--9}.
\newblock
\showeprint[arxiv]{2305.02195}
\href{https://doi.org/10.1145/3588432.3591541}{doi:\nolinkurl{10.1145/3588432.3591541}}


\bibitem[Timmer and Koenig(1995)]%
        {timmer1995generating}
\bibfield{author}{\bibinfo{person}{Jens Timmer} {and} \bibinfo{person}{Michel
  Koenig}.} \bibinfo{year}{1995}\natexlab{}.
\newblock \showarticletitle{On generating power law noise.}
\newblock \bibinfo{journal}{\emph{Astronomy and Astrophysics, v. 300, p. 707}}
  \bibinfo{volume}{300} (\bibinfo{year}{1995}), \bibinfo{pages}{707}.
\newblock


\bibitem[Tirinzoni et~al\mbox{.}(2025)]%
        {tirinzoni2025zero}
\bibfield{author}{\bibinfo{person}{Andrea Tirinzoni}, \bibinfo{person}{Ahmed
  Touati}, \bibinfo{person}{Jesse Farebrother}, \bibinfo{person}{Mateusz
  Guzek}, \bibinfo{person}{Anssi Kanervisto}, \bibinfo{person}{Yingchen Xu},
  \bibinfo{person}{Alessandro Lazaric}, {and} \bibinfo{person}{Matteo
  Pirotta}.} \bibinfo{year}{2025}\natexlab{}.
\newblock \showarticletitle{Zero-Shot Whole-Body Humanoid Control via
  Behavioral Foundation Models}. In \bibinfo{booktitle}{\emph{The Thirteenth
  International Conference on Learning Representations}}.
\newblock
\urldef\tempurl%
\url{https://openreview.net/forum?id=9sOR0nYLtz}
\showURL{%
\tempurl}


\bibitem[Touati and Ollivier(2021)]%
        {touatiLearningOneRepresentation2021}
\bibfield{author}{\bibinfo{person}{Ahmed Touati} {and} \bibinfo{person}{Yann
  Ollivier}.} \bibinfo{year}{2021}\natexlab{}.
\newblock \showarticletitle{Learning one representation to optimize all
  rewards}.
\newblock \bibinfo{journal}{\emph{Advances in Neural Information Processing
  Systems}}  \bibinfo{volume}{34} (\bibinfo{year}{2021}),
  \bibinfo{pages}{13--23}.
\newblock


\bibitem[Touati et~al\mbox{.}(2023)]%
        {touatiDoesZeroShotReinforcement2023}
\bibfield{author}{\bibinfo{person}{Ahmed Touati},
  \bibinfo{person}{J{\'e}r{\'e}my Rapin}, {and} \bibinfo{person}{Yann
  Ollivier}.} \bibinfo{year}{2023}\natexlab{}.
\newblock \showarticletitle{Does Zero-Shot Reinforcement Learning Exist?}. In
  \bibinfo{booktitle}{\emph{The Eleventh International Conference on Learning
  Representations}}.
\newblock
\urldef\tempurl%
\url{https://openreview.net/forum?id=MYEap_OcQI}
\showURL{%
\tempurl}


\bibitem[Truong et~al\mbox{.}(2024)]%
        {truong2024pdp}
\bibfield{author}{\bibinfo{person}{Takara~Everest Truong},
  \bibinfo{person}{Michael Piseno}, \bibinfo{person}{Zhaoming Xie}, {and}
  \bibinfo{person}{Karen Liu}.} \bibinfo{year}{2024}\natexlab{}.
\newblock \showarticletitle{PDP: Physics-Based Character Animation via
  Diffusion Policy}. In \bibinfo{booktitle}{\emph{SIGGRAPH Asia 2024 Conference
  Papers}}. \bibinfo{publisher}{Association for Computing Machinery},
  \bibinfo{address}{New York, NY, USA}, Article \bibinfo{articleno}{86},
  \bibinfo{numpages}{10}~pages.
\newblock
\href{https://doi.org/10.1145/3680528.3687683}{doi:\nolinkurl{10.1145/3680528.3687683}}


\bibitem[Williams(1992)]%
        {williams1992reinforce}
\bibfield{author}{\bibinfo{person}{Ronald~J Williams}.}
  \bibinfo{year}{1992}\natexlab{}.
\newblock \showarticletitle{Simple statistical gradient-following algorithms
  for connectionist reinforcement learning}.
\newblock \bibinfo{journal}{\emph{Machine learning}} \bibinfo{volume}{8},
  \bibinfo{number}{3} (\bibinfo{year}{1992}), \bibinfo{pages}{229--256}.
\newblock


\bibitem[Won et~al\mbox{.}(2020)]%
        {won2020scalable}
\bibfield{author}{\bibinfo{person}{Jungdam Won}, \bibinfo{person}{Deepak
  Gopinath}, {and} \bibinfo{person}{Jessica Hodgins}.}
  \bibinfo{year}{2020}\natexlab{}.
\newblock \showarticletitle{A scalable approach to control diverse behaviors
  for physically simulated characters}.
\newblock \bibinfo{journal}{\emph{ACM Transactions on Graphics (TOG)}}
  \bibinfo{volume}{39}, \bibinfo{number}{4} (\bibinfo{year}{2020}),
  \bibinfo{pages}{33--1}.
\newblock


\bibitem[Won et~al\mbox{.}(2022)]%
        {won2022physics}
\bibfield{author}{\bibinfo{person}{Jungdam Won}, \bibinfo{person}{Deepak
  Gopinath}, {and} \bibinfo{person}{Jessica Hodgins}.}
  \bibinfo{year}{2022}\natexlab{}.
\newblock \showarticletitle{Physics-based character controllers using
  conditional vaes}.
\newblock \bibinfo{journal}{\emph{ACM Transactions on Graphics (TOG)}}
  \bibinfo{volume}{41}, \bibinfo{number}{4} (\bibinfo{year}{2022}).
\newblock


\bibitem[Wu et~al\mbox{.}(2025)]%
        {wu2025uniphys}
\bibfield{author}{\bibinfo{person}{Yan Wu}, \bibinfo{person}{Korrawe
  Karunratanakul}, \bibinfo{person}{Zhengyi Luo}, {and} \bibinfo{person}{Siyu
  Tang}.} \bibinfo{year}{2025}\natexlab{}.
\newblock \showarticletitle{UniPhys: Unified Planner and Controller with
  Diffusion for Flexible Physics-Based Character Control}. In
  \bibinfo{booktitle}{\emph{Proceedings of the IEEE/CVF International
  Conference on Computer Vision (ICCV)}}.
\newblock


\bibitem[Xu et~al\mbox{.}(2025)]%
        {xu2025parkour}
\bibfield{author}{\bibinfo{person}{Michael Xu}, \bibinfo{person}{Yi Shi},
  \bibinfo{person}{KangKang Yin}, {and} \bibinfo{person}{Xue~Bin Peng}.}
  \bibinfo{year}{2025}\natexlab{}.
\newblock \showarticletitle{PARC: Physics-based Augmentation with Reinforcement
  Learning for Character Controllers}. In \bibinfo{booktitle}{\emph{Proceedings
  of the Special Interest Group on Computer Graphics and Interactive Techniques
  Conference Conference Papers}} \emph{(\bibinfo{series}{SIGGRAPH Conference
  Papers '25})}. \bibinfo{publisher}{Association for Computing Machinery},
  \bibinfo{address}{New York, NY, USA}, Article \bibinfo{articleno}{131},
  \bibinfo{numpages}{11}~pages.
\newblock
\showISBNx{9798400715402}
\href{https://doi.org/10.1145/3721238.3730616}{doi:\nolinkurl{10.1145/3721238.3730616}}


\bibitem[Xu and Karamouzas(2021)]%
        {xu2021gan}
\bibfield{author}{\bibinfo{person}{Pei Xu} {and} \bibinfo{person}{Ioannis
  Karamouzas}.} \bibinfo{year}{2021}\natexlab{}.
\newblock \showarticletitle{A {GAN}-Like Approach for Physics-Based Imitation
  Learning and Interactive Character Control}.
\newblock \bibinfo{journal}{\emph{Proceedings of the ACM on Computer Graphics
  and Interactive Techniques}} \bibinfo{volume}{4}, \bibinfo{number}{3}
  (\bibinfo{year}{2021}), \bibinfo{pages}{1--22}.
\newblock
\href{https://doi.org/10.1145/3480148}{doi:\nolinkurl{10.1145/3480148}}


\bibitem[Yao et~al\mbox{.}(2024)]%
        {yao2024moconvq}
\bibfield{author}{\bibinfo{person}{Heyuan Yao}, \bibinfo{person}{Zhenhua Song},
  \bibinfo{person}{Yuyang Zhou}, \bibinfo{person}{Tenglong Ao},
  \bibinfo{person}{Baoquan Chen}, {and} \bibinfo{person}{Libin Liu}.}
  \bibinfo{year}{2024}\natexlab{}.
\newblock \showarticletitle{Moconvq: Unified physics-based motion control via
  scalable discrete representations}.
\newblock \bibinfo{journal}{\emph{ACM Transactions on Graphics (TOG)}}
  \bibinfo{volume}{43}, \bibinfo{number}{4} (\bibinfo{year}{2024}),
  \bibinfo{pages}{1--21}.
\newblock


\bibitem[Yuan and Kitani(2020)]%
        {yuan2020residual}
\bibfield{author}{\bibinfo{person}{Ye Yuan} {and} \bibinfo{person}{Kris
  Kitani}.} \bibinfo{year}{2020}\natexlab{}.
\newblock \showarticletitle{Residual Force Control for Agile Human Behavior
  Imitation and Extended Motion Synthesis}. In
  \bibinfo{booktitle}{\emph{Advances in Neural Information Processing
  Systems}}.
\newblock


\bibitem[Yuksel et~al\mbox{.}(2009)]%
        {yuksel2011parameterization}
\bibfield{author}{\bibinfo{person}{Cem Yuksel}, \bibinfo{person}{Scott
  Schaefer}, {and} \bibinfo{person}{John Keyser}.}
  \bibinfo{year}{2009}\natexlab{}.
\newblock \showarticletitle{On the parameterization of Catmull-Rom curves}. In
  \bibinfo{booktitle}{\emph{2009 SIAM/ACM Joint Conference on Geometric and
  Physical Modeling}} (San Francisco, California) \emph{(\bibinfo{series}{SPM
  '09})}. \bibinfo{publisher}{Association for Computing Machinery},
  \bibinfo{address}{New York, NY, USA}, \bibinfo{pages}{47--53}.
\newblock
\showISBNx{9781605587110}
\href{https://doi.org/10.1145/1629255.1629262}{doi:\nolinkurl{10.1145/1629255.1629262}}


\bibitem[Zargarbashi et~al\mbox{.}(2025)]%
        {zargarbashi2025robotkeyframing}
\bibfield{author}{\bibinfo{person}{Fatemeh Zargarbashi}, \bibinfo{person}{Jin
  Cheng}, \bibinfo{person}{Dongho Kang}, \bibinfo{person}{Robert Sumner}, {and}
  \bibinfo{person}{Stelian Coros}.} \bibinfo{year}{2025}\natexlab{}.
\newblock \showarticletitle{RobotKeyframing: Learning Locomotion with
  High-Level Objectives via Mixture of Dense and Sparse Rewards}. In
  \bibinfo{booktitle}{\emph{Conference on Robot Learning}}
  \emph{(\bibinfo{series}{Proceedings of Machine Learning Research},
  Vol.~\bibinfo{volume}{270})}, \bibfield{editor}{\bibinfo{person}{Pulkit
  Agrawal}, \bibinfo{person}{Oliver Kroemer}, {and} \bibinfo{person}{Wolfram
  Burgard}} (Eds.). \bibinfo{publisher}{PMLR}, \bibinfo{pages}{916--932}.
\newblock
\urldef\tempurl%
\url{https://proceedings.mlr.press/v270/zargarbashi25a.html}
\showURL{%
\tempurl}


\bibitem[Zheng et~al\mbox{.}(2025)]%
        {zhengCanMISLFly2025}
\bibfield{author}{\bibinfo{person}{Chongyi Zheng}, \bibinfo{person}{Jens
  Tuyls}, \bibinfo{person}{Joanne Peng}, {and} \bibinfo{person}{Benjamin
  Eysenbach}.} \bibinfo{year}{2025}\natexlab{}.
\newblock \bibinfo{title}{Can a {{MISL Fly}}? {{Analysis}} and {{Ingredients}}
  for {{Mutual Information Skill Learning}}}.
\newblock
\showeprint[arxiv]{2412.08021}~[cs]
\href{https://doi.org/10.48550/arXiv.2412.08021}{doi:\nolinkurl{10.48550/arXiv.2412.08021}}


\end{thebibliography}

\appendix

\section{Alternative Sparse Keyframe Parametrizations}

In the sparse setting, the tracking objective is under-determined, as any latent trajectory passing through the keyframes is a valid solution. Therefore, imposing structural priors on the latent sequence is essential to effectively regularize the optimization space of the policy gradient formulation. While we demonstrate that the temporal structure provided by the correlated noise is sufficient to guide the policy toward coherent in-between motion, other valid solutions exist. To validate our design choice, we ablate this approach against alternative methods for imposing structure, such as holding the latent constant between keyframes or parameterizing the trajectory with a continuous spherical spline.

\subsection{Piecewise-Constant Latent Sequences}\label{sec:piecewise-const}

The simplest parameterization assigns each keyframe's latent to all preceding timesteps back to the previous keyframe: every timestep in segment $[t_{k-1}, t_k)$ shares the latent mean $\boldsymbol{\mu}_k$ of the upcoming keyframe $t_k$. We sample the latent sequence by drawing a single latent $\mathbf{z}_t$ per keyframe and replicating it across the segment
\begin{align}
\mathbf{z}_k &\sim \mathcal{N}(\boldsymbol{\mu}_k, \sigma^2 \mathbf{I}), \quad k = 1,\ldots,K.
\end{align}

\subsection{Spherical Spline}\label{sec:squad-spline}

To achieve smooth transitions between behaviors, we represent the latent trajectory as a continuous spherical spline using Spherical QUADrangle (SQUAD) interpolation~\citep{shoemake1985animating}, which constructs $C^1$-smooth curves on the unit hypersphere $\mathbb{S}^{d-1}$.

\paragraph{SQUAD formula.}
Given control points $\{q_i\}_{i=0}^{K-1}$ at keyframes and auxiliary inner points $\{a_i\}$ (defined below), the interpolated point at local parameter $\tau \in [0,1]$ within segment $[i, i{+}1]$ is
\begin{align}\label{eq:squad-eval}
    \mathrm{SQUAD}(\tau) &= \mathrm{Slerp}\big(\mathrm{Slerp}(q_i, q_{i+1}, \tau),\; \mathrm{Slerp}(a_i, a_{i+1}, \tau),\; h\big), 
    \\h &= 2\tau(1{-}\tau).
\end{align}

\paragraph{Auxiliary points.}
SQUAD requires auxiliary inner points $a_i$ that encode the local tangent direction at each control point, analogous to Catmull--Rom tangents in Euclidean space. These are constructed by computing the incoming and outgoing angular velocities at $q_i$, normalized by the knot intervals:
\begin{align}\label{eq:squad-velocities}
    v_i^- = \frac{\mathrm{Log}_{q_i}(q_{i-1})}{\Delta t_i^-}, \qquad
    v_i^+ = \frac{\mathrm{Log}_{q_i}(q_{i+1})}{\Delta t_i^+},
\end{align}
where $\Delta t_i^- = t_i - t_{i-1}$ and $\Delta t_i^+ = t_{i+1} - t_i$ are the knot spacings. Averaging and scaling by $-\tfrac{1}{4}$ yields the auxiliary tangent, which is then mapped back to the sphere:
\begin{align}\label{eq:squad-aux}
    a_i = \mathrm{Exp}_{q_i}\!\big(\!-\!\tfrac{1}{4}(v_i^- + v_i^+)\big).
\end{align}

\paragraph{Centripetal knot parameterization.}
The knot spacings $\Delta t_i$ directly control the tangent magnitude at each control point: a larger $\Delta t$ damps the auxiliary tangent, reducing overshoot between keyframes. We define the knot vector using centripetal parameterization:
\begin{align}
    t_i = \sum_{j < i} \|\mathrm{Log}(q_j, q_{j+1})\|^{1/2},
\end{align}
where $\|\mathrm{Log}(\cdot, \cdot)\|$ is the geodesic distance on $\mathbb{S}^{d-1}$. This assigns knot increments proportional to the square root of the arc length between consecutive control points. Segments where the latent changes rapidly receive a larger $\Delta t$, which damps the tangent and prevents oscillation---extending the well-known cusps-free property of centripetal Catmull--Rom splines~\citep{yuksel2011parameterization} to the hypersphere. Importantly, this parameterization does not alter the keyframe positions in the output sequence; it solely affects the shape of the interpolated curve between keyframes through the tangent scaling in \eqnref{eq:squad-velocities}.

\paragraph{Sampling and policy gradient.}\label{sec:squad-var}
To optimize the control points via policy gradient, we need $\log \pi(\mathbf{z}_t \mid \theta)$ for sampled trajectory points. We sample by drawing independent tangent-space perturbations $\delta_i \sim \mathcal{N}(0, \sigma_i^2 \mathbf{I}_d)$ at each control point, projecting onto the tangent plane, mapping to the sphere via $\tilde{q}_i = \mathrm{Exp}_{q_i}(\delta_i)$, and interpolating $\mathbf{z}_t = \mathrm{SQUAD}(\tilde{q}_0, \ldots, \tilde{q}_{K-1}, t)$. Because perturbations are applied at control points and propagated through the spline, every sampled trajectory is itself a spline---the function class is preserved across all rollouts.

Since SQUAD is nonlinear, the exact density of $\mathbf{z}_t$ has no closed form. Analogous to Probabilistic Movement Primitives~\cite{paraschos2013probabilistic}, but extended to the Riemannian setting, we apply a first-order linearization: for small $\sigma_i$ the manifold is locally Euclidean and the SQUAD formula reduces to a weighted linear combination of control-point perturbations. Because each auxiliary point $a_i$ depends on the neighboring keypoints $q_{i-1}, q_i, q_{i+1}$ via \eqnref{eq:squad-aux}, its variance folds into the effective influence of those keypoints. The resulting variance at parameter $\tau$ in segment $[i, i{+}1]$ (with $h = 2\tau(1{-}\tau)$) is
\begin{align}\label{eq:squad-variance}
    \sigma^2(\tau) = W_i^2\,\sigma_i^2 + W_{i+1}^2\,\sigma_{i+1}^2 + W_{i-1}^2\,\sigma_{i-1}^2 + W_{i+2}^2\,\sigma_{i+2}^2,
\end{align}
where the effective weights are
\begin{align}\label{eq:squad-eff-weights}
    W_i^2 &= (1{-}\tau)^2(1{-}h)^2 + (1{-}\tau)^2 h^2 + \tfrac{0.25^2}{(\Delta t_{i+1}^-)^2}\,\tau^2 h^2, \nonumber\\
    W_{i+1}^2 &= \tau^2(1{-}h)^2 + \tau^2 h^2 + \tfrac{0.25^2}{(\Delta t_i^+)^2}\,(1{-}\tau)^2 h^2, \nonumber\\
    W_{i-1}^2 &= \tfrac{0.25^2}{(\Delta t_i^-)^2}\,(1{-}\tau)^2 h^2, \nonumber\\
    W_{i+2}^2 &= \tfrac{0.25^2}{(\Delta t_{i+1}^+)^2}\,\tau^2 h^2.
\end{align}
The first two terms in $W_i^2$ and $W_{i+1}^2$ are the direct SLERP contributions of the segment endpoints; the remaining terms propagate through the auxiliary points. This recovers a tractable Gaussian at every timestep, enabling standard policy gradient optimization of the $K$ control points. Given the mean path $\bar{\mathbf{z}}_t = \mathrm{SQUAD}(q_0,\ldots,q_{K-1},t)$ and a sample $\mathbf{z}_t$, we compute the tangent-space residual $v_t = \mathrm{Log}_{\bar{\mathbf{z}}_t}(\mathbf{z}_t)$ and evaluate:
\begin{align}\label{eq:squad-logprob}
    \log \pi(\mathbf{z}_t \mid \theta) \approx -\frac{1}{2}\sum_{j=1}^d \left[\frac{v_{t,j}^2}{\sigma_j^2(\tau_t)} + \log\!\big(2\pi\,\sigma_j^2(\tau_t)\big)\right].
\end{align}
This approximation is exact in the limit $\sigma \to 0$ where the manifold curvature vanishes.

\section{Keyframe Extraction Heuristic}
Given a kinematic reference motion of $T$ frames, we extract keyframes using a difference-of-Gaussians (DoG) prominence heuristic on the joint kinetic energy signal.

\begin{figure}[h]
    \centering
    \includegraphics[width=\columnwidth]{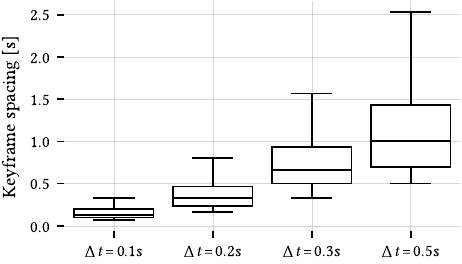}
    \caption{\textbf{Keyframe Distribution.} Distribution of keyframe spacings across the AMASS test set for varying minimum temporal spacing parameters $\Delta t$ using the DoG extraction heuristic. Box plots display the median and interquartile range (IQR), with whiskers extending to $1.5\times$ the IQR.}
    \label{fig:keyframe-spacing}
\end{figure}

Let $E(t) = \sum_j \dot{q}_j(t)^2$ denote the sum of squared joint velocities at frame $t$. We define the prominence signal as
\begin{align}
    p(t) = \left| (G_{\sigma_1} * E)(t) - (G_{\sigma_2} * E)(t) \right|,
\end{align}
where $G_\sigma$ denotes a 1D Gaussian filter with standard deviation $\sigma$, and $*$ is convolution along the time axis.
We use $\sigma_1 \approx 3.3 \cdot 10^{-5}s$ and $\sigma_2 = 2\text{s}$, making the first term match the raw energy signal very closely while $\sigma_2 = 15$ produces a smoothed baseline. 
The prominence $p(t)$ is therefore large at frames where the instantaneous energy deviates most from its local average, corresponding to rapid movements or brief pauses.
We select keyframes by applying non-maximum suppression (NMS) to the prominence signal based on a minimum temporal spacing parameter $\Delta t$. Specifically, we retain a frame $t$ if $p(t)$ is the maximum within a temporal window of size $2\Delta t$. By increasing $\Delta t \in \{0.1, 0.2, 0.3, 0.5\}$\,\SI{}{\second}, we generate keyframe sets of decreasing density. Evaluating these sets across all AMASS test clips yields median keyframe spacings of $\{0.13, 0.33, 0.67, 1.0\}$\,\SI{}{\second}, as shown in \figref{fig:keyframe-spacing}. At a standard framerate of \SI{30}{fps}, these spacings correspond to keyframe rates of $\{7.5, 3.0, 1.5, 1.0\}$~\si{\per\second}, respectively.

\section{BFM Training Details}\label{sec:app_impl-details}

Our BFM training follows the FB-CPR algorithm of \citet{tirinzoni2025zero} with the massively parallel training setup of \citet{li2025bfm}. Table~\ref{tab:hparams-diff} reports parameters that differ from these works; all other settings are identical.

\begin{table}[h]
\centering
\caption{\textbf{BFM Training Parameters}. BFM training parameters that differ from \citet{tirinzoni2025zero} and \citet{li2025bfm}.}\label{tab:hparams-diff}
\begin{tabular}{lcc}
\toprule
\textbf{Parameter} & \textbf{SMPL} & \textbf{Lima} \\
\midrule
Episode length $T$ & 300 & 500 \\
Seeding steps (random actions) & 8 & 500 \\
Fall initialization probability & 0.1 & 0.2 \\
Replay buffer size & $\approx$3M & $\approx$5M \\
Discriminator reg.\ coef.\ $\alpha_D$ & 0.01 & 0.05 \\
Auxiliary reward reg.\ coef.\ $\alpha_R$ & 0.001 & 0.5 \\
Rollout noise $\sigma \sim \mathcal{U}(\sigma_\text{min}, \sigma_\text{max})$~\cite{seoFastTD3SimpleFast2025} & \multicolumn{2}{c}{$(0.001,\, 0.3)$} \\
Actor training std & \multicolumn{2}{c}{0.005} \\
Std clip & \multicolumn{2}{c}{0.015} \\
Learning rates ($F$, $\pi$, $Q_R$) & \multicolumn{2}{c}{$3\times10^{-4}$} \\
\bottomrule
\end{tabular}
\end{table}

Following \citet{seoFastTD3SimpleFast2025}, we sample the rollout exploration noise $\sigma$ independently per environment from a uniform distribution $\mathcal{U}(\sigma_\text{min}, \sigma_\text{max})$, refreshing on episode reset. This diversity of exploration scales improves off-policy learning without requiring a fixed noise schedule.

\paragraph{Prioritized motion sampling.}
To address the inherent imbalance in motion difficulty within the AMASS dataset, we partition all sequences into \SI{10}{\second} chunks and dynamically re-weight their sampling frequencies during training. We periodically evaluate the policy's tracking performance on these chunks to prioritize the exploration of more difficult clips. For each chunk, we roll out the policy and compute the average tracking error in the root body frame by stacking all rigid-body positions
\begin{align}
    e = \frac{1}{T}\sum_{t=1}^{T} \left\| 
    \begin{bmatrix} \mathbf{p}_t^1 \\ \vdots \\ \mathbf{p}_t^{|\mathcal{J}|} \end{bmatrix} - 
    \begin{bmatrix} \hat{\mathbf{p}}_t^1 \\ \vdots \\ \hat{\mathbf{p}}_t^{|\mathcal{J}|} \end{bmatrix} 
    \right\|_2,
\end{align}
where $\mathbf{p}_t^i, \hat{\mathbf{p}}_t^i \in \mathbb{R}^3$ denote the $i$-th rigid-body position of the simulated and reference trajectories, respectively, for a total of $|\mathcal{J}|$ bodies. 

The resulting error $e$ dictates the updated sampling probability for each chunk, with distinct schedules and functions applied depending on the target character. For the SMPL humanoid, we re-weight the dataset every 100k gradient steps using histogram equalization. Specifically, chunk errors are sorted into 8 uniform bins across the $[20, 100]$\SI{}{\centi\meter} range, and each clip is assigned a weight inversely proportional to its bin's total count. This guarantees that the agent samples uniformly across all difficulty levels. For the Lima character, we instead re-weight the dataset every 400k gradient steps using an exponential upweighting scheme
\begin{align}
    w = 2^{\,6\,\cdot\,\text{clamp}(e,\, 200/3,\, 200)\,/\,200}.
\end{align}
This formulation assigns exponentially higher sampling probabilities to clips with larger tracking errors, aggressively directing the optimization toward the most challenging motions.

\subsection{Auxiliary Regularization Rewards}\label{sec:lima-reward}
Following BFM-Zero~\cite{li2025bfm}, we apply during the BFM training an auxiliary regularization reward for humanoid robot control listed in \tabref{tab:aux-rewards}. %

\begin{table}[h]
\centering
\caption{\textbf{Auxiliary Regularization Reward}. Auxiliary reward terms for BFM training.}\label{tab:aux-rewards}
\begin{tabular}{lcc}
\toprule
\textbf{Term} & \textbf{SMPL} & \textbf{Lima} \\
\midrule
Torques (legs) & \multirow{2}{*}{$-2.5\times10^{-5}$} & $-1.0\times10^{-3}$ \\
Torques (arms and neck) & & $-1.0\times10^{-2}$ \\
Acceleration (legs) & \multirow{2}{*}{$-2.5\times10^{-6}$} & $-2.5\times10^{-6}$ \\
Acceleration (arms and neck) & & $-2.5\times10^{-5}$ \\
Action rate (legs) & \multirow{2}{*}{$-1.0\times10^{-2}$} & $-30$ \\
Action rate (arms and neck) & & $-30$ \\
Action 2nd derivative (legs) & \multirow{2}{*}{$-1.0\times10^{-2}$} & $-20$ \\
Action 2nd derivative (arms and neck) & & $-20$ \\
Self-contact & --- & $-1.0$ \\
Hard-stop impact (CBF) & --- & $-1.0$ \\
\bottomrule
\end{tabular}
\end{table}

\end{document}